\PassOptionsToPackage{table}{xcolor}
\documentclass[lettersize,journal]{IEEEtran}

\usepackage{tikz}

\newcommand\submittedtext{
  \footnotesize This work has been submitted to the IEEE for possible publication. Copyright may be transferred without notice, after which this version may no longer be accessible.}

\newcommand\submittednotice{
\begin{tikzpicture}[remember picture,overlay]
\node[anchor=south,yshift=10pt] at (current page.south) {\fbox{\parbox{\dimexpr\textwidth-\fboxsep-\fboxrule\relax}{\submittedtext}}};
\end{tikzpicture}
}

\usepackage{amsmath}
\usepackage{amsfonts}
\usepackage{amssymb}
\usepackage{symbols}
\usepackage{xcolor}
\usepackage{colors}
\usepackage{enumitem}

\usepackage{amsmath,amsfonts,bm}

\def\1{\bm{1}}

\def\sp{space}

\def\ve{{\bm{e}}}
\def\vf{{\bm{f}}}
\def\vg{{\bm{g}}}

\def\vi{{\bm{i}}}

\def\vx{{\bm{x}}}
\def\vy{{\bm{y}}}

\def\mK{{\bm{K}}}

\def\mS{{\bm{S}}}
\def\mT{{\bm{T}}}

\DeclareMathAlphabet{\mathsfit}{\encodingdefault}{\sfdefault}{m}{sl}
\SetMathAlphabet{\mathsfit}{bold}{\encodingdefault}{\sfdefault}{bx}{n}

\def\gT{{\mathcal{T}}}

\def\sR{{\mathbb{R}}}

\usepackage[ruled,vlined]{algorithm2e}
\SetAlgorithmName{Alg.}{}{}
\SetAlFnt{\small}
\SetKwInput{KwInput}{Input}
\SetKwInput{KwOutput}{Output}
\SetKwInput{KwParameters}{Parameters}
\SetKwComment{Comment}{\# }{}

\usepackage{ctable} %
\usepackage{multirow}
\usepackage{makecell}
\usepackage{epstopdf}
\usepackage[numbers]{natbib}

\definecolor{bg_blue}{HTML}{e6efff}

\usepackage{pifont}
\newcommand{\cmark}{\ding{51}}
\newcommand{\xmark}{\ding{55}}

\ifCLASSINFOpdf
\else
\fi

 \usepackage[caption=false,font=footnotesize]{subfig}

\usepackage[pagebackref=true,
            breaklinks=true,
            colorlinks,
            linkcolor=mycitecolor,
            anchorcolor=mycitecolor,
            citecolor=mycitecolor,
            bookmarks=false,
            urlcolor=magenta,
            ]{hyperref}

\begin{document}
\title{GIST: Towards Photorealistic Style Transfer\\via Multiscale Geometric Representations}

\author{Renan A. Rojas-Gomez,~\IEEEmembership{Member,~IEEE;}
        Minh N. Do,~\IEEEmembership{Fellow,~IEEE}
\thanks{The authors are with the Electrical and Computer Engineering Department, University of Illinois Urbana-Champaign. Contact information: renanar2@illinois.edu, minhdo@illinois.edu. This work is supported in part by PPG Industries, the Jump ARCHES Endowment through the Health Care Engineering Systems Center, and GlaxoSmithKline (GSK) R\&D Ltd.}%
}

\maketitle
\submittednotice
\begin{abstract}
State-of-the-art Style Transfer methods often leverage pre-trained encoders optimized for discriminative tasks, which may not be ideal for image synthesis. This can result in significant artifacts and loss of photorealism. Motivated by the ability of multiscale geometric image representations to capture fine-grained details and global structure, we propose \textit{GIST: Geometric-based Image Style Transfer}, a novel Style Transfer technique that exploits the geometric properties of content and style images. GIST replaces the standard Neural Style Transfer autoencoding framework with a multiscale image expansion, preserving scene details without the need for post-processing or training. Our method matches multiresolution and multidirectional representations such as Wavelets and Contourlets by solving an optimal transport problem, leading to an efficient texture transferring. Experiments show that GIST is on-par or outperforms recent photorealistic Style Transfer approaches while significantly reducing the processing time with no model training. Project website: \url{https://github.com/renanrojasg/gist}.
\end{abstract}

\begin{IEEEkeywords}
Example-based Style Transfer, Multiresolution Image Representation, Wavelet Transform, Multidirectional Image Representation, Contourlet Transform.
\end{IEEEkeywords}

\IEEEpeerreviewmaketitle
\section{Introduction}
\begin{figure}[t]
\vspace{-0.05cm}
\noindent\fcolorbox{white}{white}{\begin{minipage}{0.12\columnwidth}
\centering {\scriptsize{\vspace{0.025cm}\textbf{Refs.}}\\Content\\[-0.15cm]Style\vspace{-0.075cm}}
\end{minipage}}\noindent\fcolorbox{white}{white}{\begin{minipage}{0.2575\columnwidth}
\centering {\scriptsize{\textbf{WCT$^{2}$ (\texttt{Sum})}}\\Parameters: $3.5$M\\[-0.15cm]Inference time: $0.37$s}
\end{minipage}}\noindent\fcolorbox{white}{white}{\begin{minipage}{0.2575\columnwidth}
\centering {\scriptsize{\textbf{WCT$^{2}$ (\texttt{Concat})}}\\Parameters: $6.6$M\\[-0.15cm]Inference time: $0.45$s}
\end{minipage}}\noindent\fcolorbox{white}{white}{\begin{minipage}{0.2575\columnwidth}
\centering {\scriptsize{\textcolor{mybb}{\textbf{GIST (Ours)}\\Parameters: None\\[-0.15cm]Inference time: $0.1$s}}}
\vspace{-0.05cm}
\end{minipage}}

\includegraphics[width=\columnwidth]{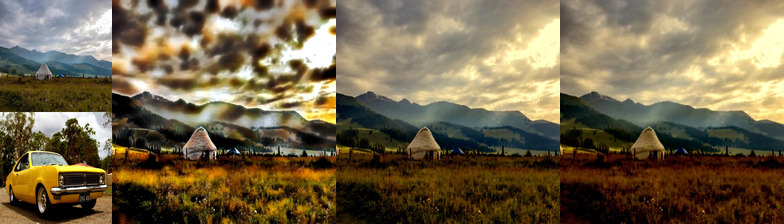}

\includegraphics[width=\columnwidth]{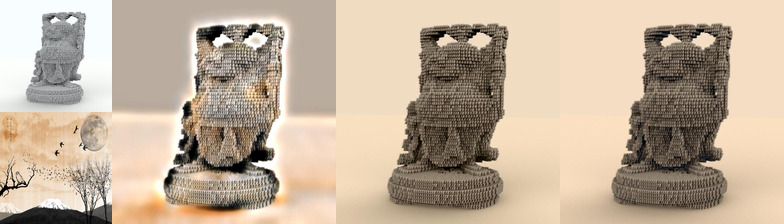}

\includegraphics[width=\columnwidth]{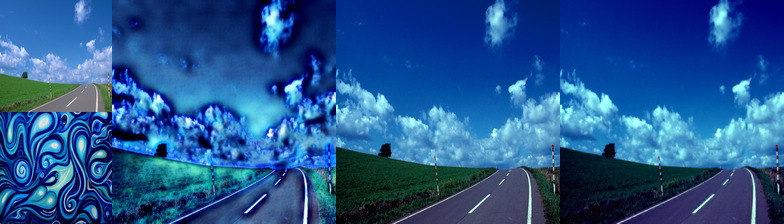}
\vspace{-1.5\baselineskip}

\caption{\textbf{Photorealistic Style Transfer via geometric image representations.} We propose GIST, a Geometric-based Image Style Transfer technique that aligns multiscale representations such as Wavelets and Contourlets to efficiently transfer style from arbitrary images. Our method achieves improved or on-par performance to deep learning methods like WCT$^{2}$ in content and texture preservation without the need for training or extensive computations.}
\label{fig:st_photo_contrast}
\vspace{-0.5\baselineskip}
\end{figure}

Style is defined as a class of images sharing common statistical properties \cite{zhu_2000_exploring}. Traditional Texture Synthesis and Style Transfer methods use pre-selected statistics to quantify the similarity between images \cite{zhu_2000_exploring, portilla_2000_parametric, heeger_1995_pyramid}. For instance, by iteratively updating an image, often initialized as noise, to match target statistics, Style Transfer techniques generate a novel view that combines the visual features of a style image with the objects of a content image, as shown in \figref{fig:st_photo_contrast}.

More recently, Deep Neural Networks have emerged as powerful tools for characterizing style \cite{gatys_2017_controlling}. Extracting feature maps from a pre-trained network allow these methods to capture high-level content and style representations, enabling the synthesis of high-quality images. By leveraging their representational power, deep learning techniques outperform traditional methods both in terms of quality and efficiency.

While early Neural Style Transfer methods relied on an iterative optimization approach to preserve content objects while imposing the style's appearance, recent techniques have adopted an \textit{encode-align-decode} approach \cite{li_2017_universal}. This involves training a decoder to invert content representations that have been aligned with a style reference. This approach offers more efficient stylization and enables the use of arbitrary content and style images, albeit with slightly degraded texture synthesis \cite{johnson_2016_perceptual}. Despite its widespread use in artistic applications, stylized images often exhibit an unnatural appearance due to information loss during the feature extraction process, limiting its use in applications requiring a natural, \textit{photorealistic} appearance.

A significant body of work has been dedicated to generating photorealistic stylization, \ie, creating stylized images that maintain a natural appearance. Most methods enforce photorealism by imposing priors on the output pixel domain \cite{luan_2017_deep,mechrez_2017_photorealistic}.
However, recent approaches explore techniques that operate directly in the latent space \cite{yoo_2019_photorealistic, an_2020_ultrafast}. While prior-based methods overpenalize the output, degrading image quality and incurring high computational costs, latent-space methods enforce natural appearance by utilizing specialized feature alignment techniques or adjusting the network architecture to minimize the distortions caused by the loss of information during feature extraction.
Nevertheless, as all these methods rely on a pre-trained classifier, image reconstruction remains suboptimal, leading to visual aberrations.

\begin{figure*}[t]
\centering
\includegraphics[width=\textwidth]{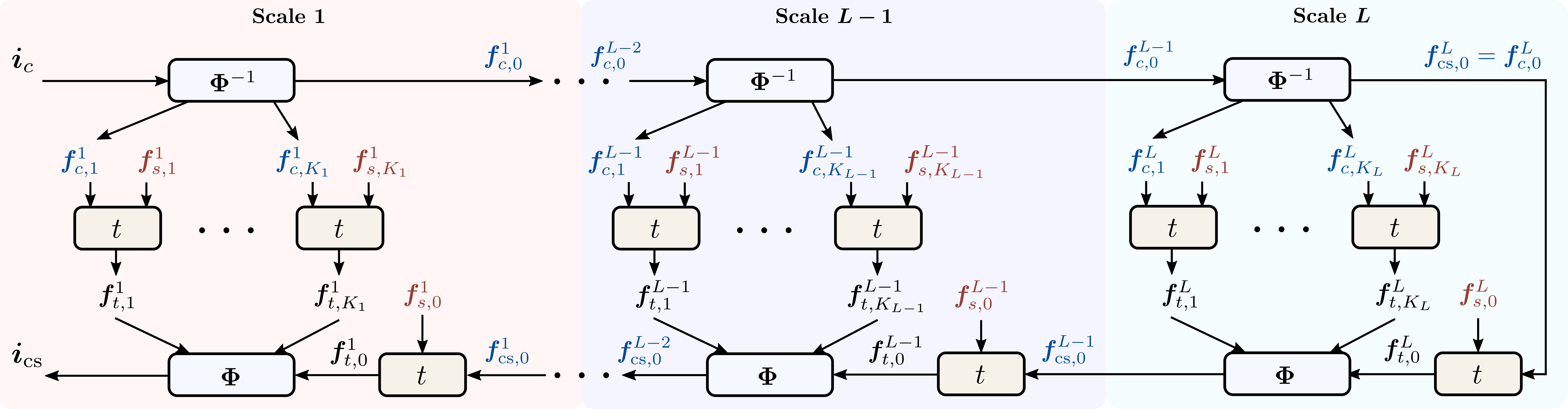}
\vspace{-0.6cm}
\caption{\label{fig:gist_diag}\textbf{GIST: Style Transfer using multiscale geometric representations.} To create a stylized image $\vi_{\text{cs}}$, GIST progressively aligns the \textcolor{contentblue}{content from $\vf_{c}$} subbands and the \textcolor{stylered}{style from $\vf_{s}$} subbands from coarse to fine resolution using an optimal transport map $t$. This ensures the preservation of content attributes while incorporating the perceptual properties of the style image. GIST can handle general geometric image representations such as Wavelets and Contourlets.}
\vspace{-0.1cm}
\end{figure*}

To overcome these limitations, we propose \textit{GIST: Geometry-based Image Style Transfer}, a novel algorithm for photorealistic Style Transfer. Inspired by the properties of multiscale geometric image expansions in preserving both global and local image details, we replace the traditional learn-based autoencoding approach with a multiscale expansion \cite{daubechies_1992_ten, do_2005_contourlet, candes_2006_fast}. This approach allows to maintain image content without requiring pixel-level regularization or training. GIST aligns style via a relaxed feature matching process formulated as an optimal transport problem. By assuming a Gaussian distribution in the latent space, we minimize the \textit{Wasserstein-$2$ distance} using a closed-form solution, leading to fast and faithful Style Transfer. \textbf{Our contributions are:}
\begin{itemize}[leftmargin=*]
    \item We enforce style and content preservation by replacing the autoencoder used in Neural Style Transfer with multiresolution and multidirectional representations. This eliminates the need for heuristic regularization or learned image features.
    \item We efficiently align content and style by matching subband distributions using the Wasserstein-2 distance. Under a Gaussian approximation, we obtain a closed-form solution for the optimal transport map, enabling an efficient style matching without compromising quality.
    \item We demonstrate empirically that our stylization framework supports multiple image expansions, including Wavelets \cite{daubechies_1992_ten} and Contourlets \cite{do_2005_contourlet}, offering greater flexibility in image representation and control over the stylization strength.
\end{itemize}

\section{Related Work}
\label{sec:related_work}
\subsection{Texture Synthesis via Multiscale Representations}
Seminal work in image synthesis explored the use of multiscale representations to describe texture. These include the pyramid-based approach of \citet{heeger_1995_pyramid}, decomposing texture into multiple frequency bands and synthesizing them independently via statistical models, the patch-based sampling by \citet{efros_1999_texture}, the use of pyramid decomposition and Markov random fields by \citet{wei_2000_fast}, and the steerable pyramid decomposition by \citet{portilla_2000_parametric}. Unlike these optimization-based methods, GIST leverages multi-scale expansions to impose style over content by extracting representations, matching their distributions, and synthesizing a stylized image in a single pass.

More recent work by \citet{fan_2003_wavelet} leverages Wavelets for texture characterization, demonstrating the effectiveness of complex Wavelet features for tasks such as classification, segmentation, and synthesis. While their generative approach relies on Hidden Markov Models to capture joint subband statistics, GIST matches the style of arbitrary images by solving a relaxed optimal transport problem. \citet{do_2002_wavelet} proposed a Wavelet-based method for texture classification and generation by parameterizing Wavelet subbands with a generalized Gaussian distribution, improving generation quality and efficiency. Our work aligns with this approach by matching subband distributions, but GIST exploits the closed-form solution of optimal transport under Gaussianity for an efficient subband matching, without being limited to Wavelets. Recent work by \citet{brochard_2022_generalized} uses nonlinear Wavelets and phase harmonics for Texture Synthesis in an optimization-based manner, capturing complex shapes at multiple orientations. While GIST also utilizes multiscale and multidirectional representations to capture complex shapes, we align content and style subbands independently in a single pass by matching low-order moments in a closed-form fashion.

\subsection{Neural Style Transfer}
Deep learning models have demonstrated significant potential in image generation and Style Transfer tasks. Recent advancements have explored measuring texture similarity between images by analyzing their representations within the latent space of pre-trained Convolutional Neural Networks (CNNs) \cite{heitz_2021_sliced,yoo_2019_photorealistic,gatys_2017_controlling,wang_2020_diversified}. For instance, prominent image classifiers like VGG-19 \cite{simonyan_2015_very} are employed to extract image features. Subsequently, low-order moments of these feature distributions are utilized as texture descriptors \cite{li_2017_universal, huang_2017_arbitrary}. By matching these feature statistics iteratively \cite{gatys_2016_image} or through closed-form solutions \cite{kessy_2018_optimal}, CNN-based techniques have achieved state-of-the-art performance in Style Transfer and Texture Synthesis. In contrast, GIST leverages geometric multiscale representations for photorealistic stylization. Our approach eliminates the need for training neural networks or relying on pre-trained models.

Conventional CNN-based Style Transfer methods, which primarily target \emph{artistic} applications, rely on iterative optimization \cite{heitz_2021_sliced,gatys_2016_image,gatys_2017_controlling} or autoencoding approaches \cite{li_2017_universal,wang_2020_diversified, huang_2017_arbitrary}. While these techniques offer flexibility, they are computationally expensive or rely on training one or more image decoders. In contrast to these techniques, GIST leverages multiscale linear synthesis and analysis linear operators for efficient image matching and reconstruction.

To enhance the \emph{natural appearance} of stylized images, various techniques have been proposed, including post-processing with edge-preserving priors \cite{luan_2017_deep,li_2018_closed,mechrez_2017_photorealistic} and refining the model architecture or operating directly on the CNN latent space \cite{yoo_2019_photorealistic,an_2020_ultrafast, gatys_2017_controlling, qiao_2021_efficient}. However, these methods can lead to suboptimal results or increased computational complexity. Conversely, GIST leverages the perfect reconstruction property of critically and oversampled image expansions, enabling the preservation of natural image structure without the need for additional regularization or prior information.

Alternative Style Transfer methods \cite{wang_2023_rethinking} use watermarking techniques to impose fine texture details, such as brushstrokes, by computing the Edge Tangent Flow \cite{kang_2007_coherent} of the style reference and incorporating it in the RGB space prior to Neural Style Transfer. While GIST achieves photorealism through multiscale representations, it also emphasizes stylistic shapes towards a more artistic output. We achieve this by progressively fusing the detail subbands of the style Edge Tangent Flow with those of the content image.

\section{Preliminaries}
\label{sec:preliminaries}
\subsection{The Wavelet Transform}
\label{sec:prelim_wavelets}
In the discrete domain, the Wavelet Transform \cite{daubechies_1992_ten, mallat_1999_wavelet} is a signal processing technique that decomposes an image into a set of coefficients representing its features at different scales and orientations. Wavelets provide a computationally efficient way to analyze details at multiple resolutions and positions.

The Wavelet transform is comprised by synthesis and analysis linear operators. The analysis operator decomposes an image into a set of Wavelet basis functions, which are localized in space and frequency, making them well-suited for analyzing transient signals. The synthesis operator consists of the reverse process, reconstructing an image from its Wavelet coefficients.

A multiscale Wavelet transform can be computed efficiently using a recursive algorithm. This involves iteratively decomposing an image into lower-resolution approximation and detail coefficients at each scale. Without loss of generality, the multiscale Wavelet representation of an image $\vi \in \mathbb{R}^{C\times W\times H}$ at the $l$-th scale can be expressed as:
\begin{align}
    \begin{pmatrix} \vf^{l}_{a} & \vf^{l}_{v} & \vf^{l}_{h} & \vf^{l}_{d}\end{pmatrix}=&\ \bm{\Phi_{w}}^{-1}\vf^{l-1}_{a}
\end{align}
where $\bm{\Phi_{w}}^{-1}$ corresponds to the Wavelet analysis operator and subindices $\{a, v, h, d\}$ denote the approximation, vertical, horizontal and diagonal subbands, respectively. While the approximation subband is comprised of the low-frequency image components at the $l$-th scale, the remaining three subbands represent high-frequency components along horizontal, vertical and diagonal directions. Here we assume that $\vf_{a}^{0}=\vi$.

Given Wavelet subbands at the $l$-th scale, the approximation component at scale $l-1$ (one finer scale) can be expressed as
\begin{align}
    \vf^{l-1}_{a}=&\ \bm{\Phi_{w}} \begin{pmatrix} \vf^{l}_{a} & \vf^{l}_{v} & \vf^{l}_{h} & \vf^{l}_{d}\end{pmatrix}
\end{align}
where $\bm{\Phi_{w}}$ corresponds to the Wavelets synthesis operator. This leads to a recursive form that allows recovering the original image as $\vi = \bm{\Phi_{w}}\begin{pmatrix} \vf^{1}_{a} & \vf^{1}_{v} & \vf^{1}_{h} & \vf^{1}_{d}\end{pmatrix}$.

{\bf \noindent Decimated Wavelet Transform.} The standard representation in the Wavelet space corresponds to the Decimated Wavelet Transform (DWT) \cite{daubechies_1992_ten}, which provides a critically-sampled multiscale image expansion. This means that the number of output coefficients is equal to the number of input samples, ensuring efficient representation without redundancy. DWT's multiscale analysis involves recursive filtering and downsampling, progressively reducing the resolution of the subbands.
\begin{align}
    \vf^{\text{DWT}, l}_{k} \in \mathbb{R}^{C \times W/2^{l} \times H/2^{l}},\ k\in \{a,v,h,d\}
\end{align}

The synthesis process reverses this operation, upsampling and filtering the coefficients to reconstruct the original image.

{\bf \noindent Stationary Wavelet Transform.} An alternative representation corresponds to the Stationary Wavelet Transform (SWT) \cite{nason_1995_stationary}, which addresses a limitation of the DWT: its lack of translation invariance. SWT overcomes this limitation by employing an oversampled representation. Unlike DWT, the stationary or undecimated version applies the same filterbank to the image at each scale without downsampling. This results in a redundant representation where the number of coefficients at each scale is the same as the number of input samples.
\begin{align}
    \vf^{\text{SWT}, l}_{k} \in \mathbb{R}^{C \times W \times H},\ k\in \{a,v,h,d\}
\end{align}

While Wavelets decompose images in three main directions, they are inefficient at representing directional features like edges and curves. An alternative representation corresponds to a finer directional image expansion known as Contourlets.

\subsection{The Contourlet Transform}
The Contourlet transform \cite{do_2005_contourlet, lu_2003_crisp} is a representation technique that offers a multidirectional and multiscale analysis. It overcomes the limitations of Wavelets, which struggle to represent directional features like edges and curves effectively.

Contourlets rely on a two-stage filtering process. First, a Laplacian Pyramid \cite{burt_1987_laplacian} is used to decompose an image into different scales. Next, a directional filterbank \cite{lu_2007_multidimensional} is applied to extract features at each scale, capturing information at multiple orientations. This combination of multiscale and multidirectional decomposition results in an oversampled representation that accurately describes complex image structures.

Similarly to Wavelets, the Contourlet Transform is comprised of synthesis and analysis linear operators. At the $l$-th scale, the analysis operator corresponds to a Laplacian Pyramid decomposition followed by a directional filterbank comprised by $K_{l}$ directional filters. This leads to the following multiscale multidirectional representation
\begin{align}
    \begin{pmatrix} \vf^{l}_{a} & \vf^{l}_{d,1} & \dots & \vf^{l}_{d, K_{l}}\end{pmatrix}=&\ \bm{\Phi_{c}}^{-1}\vf^{l-1}_{a}
\end{align}
where $\vf^{l}_{a}$ denotes the approximation subband, $(\vf^{l}_{d, k})_{k=1}^{K_{l}}$ the directional subbands and $\bm{\Phi_{c}}^{-1}$ the analysis operator.

The Contourlet synthesis process involves reconstructing the original image from its contourlet coefficients. This is achieved by reversing the analysis process, \ie, applying an inverse directional filterbank to recover the corresponding Laplacian pyramid representation at scale $l$, followed by an inverse Laplacian pyramid step to reconstruct the approximation subband at scale $l-1$ via the synthesis operator $\bm{\Phi_{c}}$
\begin{align}
    \vf^{l-1}_{a}=&\ \bm{\Phi_{c}} \begin{pmatrix} \vf^{l}_{a} & \vf^{l}_{d,1} & \dots & \vf^{l}_{d,K_{l}}\end{pmatrix}
\end{align}

This leads to a recursive form that allows recovering the original image as $\vi= \bm{\Phi_{c}} \begin{pmatrix} \vf^{1}_{a} & \vf^{1}_{d,1} & \dots & \vf^{1}_{d,K_{l}}\end{pmatrix}$.

While Wavelets and Contourlets differ in their specific constructions, they can be seen as special cases of a general framework of multiscale representations. Building on this, we propose a general multiscale encoder-decoder architecture for photorealistic Style Transfer, as detailed in Section \ref{sec:proposed_method}.

\subsection{Matching Distributions via Optimal Transport}
Optimal transport \cite{villani_2003_topics} provides a framework for measuring the distance between probability distributions by computing the minimum cost required to transform one distribution into another. Given probability spaces $(\mathcal{X}, \mu)$ and $(\mathcal{Y}, \nu)$, optimal transport seeks to find the most efficient way to transport mass from a source distribution $\mu$ to a target distribution $\nu$.

Given a cost function $\mathcal{F}:\mathcal{X}\times \mathcal{Y}\mapsto \mathbb{R}_{+}$, the optimal transport problem aims to find the optimal map $\mT:\mathcal{X}\mapsto \mathcal{Y}$ that minimizes the total transport cost $\int_{\mathcal{X}}\mathcal{F}\big(x,\mT(x)\big)d\mu(x)$ among all valid transport maps. A valid transport map $\mT$ must satisfy the pushforward condition $\nu(\mathcal{A})= \mu\big(\mT^{-1}(\mathcal{A})\big)$ for all measurable sets $\mathcal{A} \subset \mathcal{Y}$.

\noindent{\bf Closed-form Solution under Gaussianity.}
Let $\eta_{\mathcal{X}}$ and $\eta_{\mathcal{Y}}$ be Gaussian measures on $\mathcal{X}$ and $\mathcal{Y}$ with mean vectors $m_{\mathcal{X}}, m_{\mathcal{Y}}$ and covariance matrices $\Sigma_{\mathcal{X}}$ and $\Sigma_{\mathcal{Y}}$, respectively. Under the squared Euclidean cost function $\mathcal{F}(x,y)= \|x-y\|_{2}^{2}$, the optimal transport problem between $\eta_{\mathcal{X}}$ and $\eta_{\mathcal{Y}}$ reduces to the computation of the squared Wasserstein-2 distance $W_{2}^{2}$, which has a closed-form solution \cite{takatsu_2010_wasserstein}.
\begin{align}
    W_{2}^{2}(\eta_{\mathcal{X}},\eta_{\mathcal{Y}})=&\ \|m_{\mathcal{X}}-m_{\mathcal{Y}}\|_{2}^{2}+\mathcal{B}^{2}(\Sigma_{\mathcal{X}},\Sigma_{\mathcal{Y}})
\end{align}
where $\mathcal{B}^{2}$ corresponds to the \textit{Bures} distance \cite{bhatia_2019_bures}
\begin{align}
    \mathcal{B}^{2}(\Sigma_{\mathcal{X}},\Sigma_{\mathcal{Y}})=&\ \text{Tr}\big(\Sigma_{\mathcal{X}}+\Sigma_{\mathcal{Y}}-2\big(\Sigma_{\mathcal{Y}}^{1/2}\Sigma_{\mathcal{X}}\Sigma_{\mathcal{Y}}^{1/2}\big)^{1/2}\big)
\end{align}
and $\text{Tr}$ is the trace operator. Based on the Bures gradient, the optimal transport map has also a closed form solution
\begin{align}
    \mT^{\star}(x)=&\ \Sigma_{\mathcal{X}}^{-1/2}\big(\Sigma_{\mathcal{X}}^{1/2}\Sigma_{\mathcal{Y}}\Sigma_{\mathcal{X}}^{1/2}\big)^{1/2}\Sigma_{\mathcal{X}}^{-1/2}\bar{x}+m_{\mathcal{Y}}
\end{align}
for $\bar{x}=x-m_{\mathcal{X}}, x\in \mathcal{X}$.

Following this, given the probability measures $\mu$ and $\nu$ on spaces $\mathcal{X}$ and $\mathcal{Y}$, respectively, with first and second moments given by $(m_{\mathcal{X}}, \Sigma_{\mathcal{X}})$ and $(m_{\mathcal{Y}}, \Sigma_{\mathcal{Y}})$, the Wasserstein-2 distance $W_{2}^{2}(\mu,\nu)$ is lower-bounded by the Wasserstein-2 distance between the Gaussian measures $\eta_{\mathcal{X}}$ and $\eta_{\mathcal{Y}}$, \ie, $W_{2}^{2}(\mu,\nu)\geq W_{2}^{2}(\eta_{\mathcal{X}},\eta_{\mathcal{Y}})$.

The Gaussian lower bound simplifies optimal transport by reducing it to a problem of matching first and second-order statistical moments, analogous to the use of Gram loss in Neural Style Transfer \cite{mroueh_2020_wasserstein, kessy_2018_optimal, li_2017_universal, gatys_2016_image}. Our proposed method leverages this towards an efficient photorealistic Style Transfer algorithm based on multiscale representation matching.

\section{Proposed Method}
\label{sec:proposed_method}

\begin{algorithm}[t]
\caption{GIST: Geometric-based Image Style Transfer}
\label{alg:gist}
\KwInput{content $\bm{i}_{c}$, style $\bm{i}_{s}$, scales $L$, directions $(K_{l})_{l=1}^{L}$.}
\KwOutput{stylized image $\bm i_{\text{cs}}$.}
\BlankLine
\Comment{Initialize finest scale.}
$\vf_{c,0}^{0}\gets \bm{i}_{c}$\;
\BlankLine
$\vf_{s,0}^{0}\gets \bm{i}_{s}$\;
\BlankLine\BlankLine
\Comment{Get multiscale representations, Eq. \eqref{eq:analysis}.}
\For{$l=1$ to $L$}{
    $\begin{pmatrix}
    \vf_{c,0}^{l} & \dots & \vf_{c,K_{l}}^{l}
    \end{pmatrix}\gets \bm{\Phi}^{-1} \vf_{c,0}^{l-1}$\;
     \BlankLine
    $\begin{pmatrix}
    \vf_{s,0}^{l} & \dots & \vf_{s,K_{l}}^{l}
    \end{pmatrix}\gets \bm{\Phi}^{-1} \vf_{s,0}^{l-1}$\;
}
\BlankLine
\Comment{Initialize coarsest approx., Eq. \eqref{eq:decoder_initialization}.}
$\vf_{\text{cs},0}^{L}\gets \vf_{c,0}^{L}$
\BlankLine\BlankLine
\Comment{Coarse-to-fine alignment.}
\For{$l=L$ down to $1$}{
    \Comment{Build stylized rep., Eq. \eqref{eq:stylized_representation}.}
    $\vf^{l}_{\text{cs}}\gets \begin{pmatrix}
        \vf^{l}_{\text{cs},0} & \vf^{l}_{c,1} & \dots & \vf^{l}_{c,K_{l}}
     \end{pmatrix}$;
    \BlankLine\BlankLine
    \Comment{Align subbands, Eq. \eqref{eq:representation_alignment}.}
    $\vf^{l}_{t}\gets \gT(\vf^{l}_{\text{cs}}, \vf^{l}_{s})$\;
    \BlankLine\BlankLine
    \Comment{Reconstruct finer approx., Eq. \eqref{eq:finer_approximation}.}
    $\vf^{l-1}_{\text{cs},0}\gets \bm{\Phi} \vf^{l}_{t}$\;
}
\BlankLine
\Comment{Compute stylized image, Eq. \eqref{eq:stylized_image}.}
$\bm{i}_{\text{cs}}\gets \vf^{0}_{\text{cs}, 0}$
\end{algorithm}

GIST utilizes multiscale geometric representations to extract semantic and textural information from content and style images. At each scale, sparse image representations are extracted and reconstructed through the application of analysis and synthesis operators, denoted as $\bm{\Phi}^{-1}$ and $\bm{\Phi}$, respectively.

By decomposing content and style images into subbands and subsequently reconstructing them, we achieve fine-grained control over texture at various resolutions, enabling seamless fusion into stylized representations. These representations are then mapped back to the pixel domain without any loss of content detail. Importantly, the synthesis and analysis operators of multiscale geometric representations promote an accurate image reconstruction without requiring training.

\subsection{Style Transfer using Geometric Representations}
Our method incorporates three essential components: (i) a \textit{multiresolution encoder} for extracting content and style representations, (ii) a \textit{subband matching} technique for aligning multiscale representations by matching their distributions, and (iii) a \textit{multiresolution decoder} for inverting the aligned representations and generating the stylized image.

\noindent{\textbf{Multiresolution Encoder.}}
In contrast to pre-trained models used in conventional Neural Style Transfer, we employ geometric representations such as Wavelet and Contourlet coefficients to encode content and style information. This provides multiresolution image representations computed via linear synthesis and analysis operators, resulting in an efficient stylization process while preserving fine-grained image details.

Let $\vi_{c}\in \sR^{C \times W_{c} \times H_{c}}$ and $\vi_{s}\in \sR^{C \times H_{s} \times W_{s}}$ denote a pair of content and style images, respectively. Assuming a geometric image expansion with $L$ scales and $K_{l}$ directions per scale for $l \in \{1,\dots,L\}$, we extract content and style multiscale representations $(\bm{f}^{l}_{c}, \bm{f}^{l}_{s})_{l=1}^{L}$
\begin{align}
    \label{eq:analysis}
    \vf^{l}_{c}=& \begin{pmatrix}
      \vf^{l}_{c,0} & \dots & \vf^{l}_{c,K_{l}}
                  \end{pmatrix}=\ \bm{\Phi}^{-1}\vf^{l-1}_{c}\\
    \vf^{l}_{s}=& \begin{pmatrix}
      \vf^{l}_{s,0} & \dots & \vf^{l}_{s,K_{l}}
                  \end{pmatrix}=\ \bm{\Phi}^{-1}\vf^{l-1}_{s}
\end{align}
where $\bm{\Phi}^{-1}$ is the analysis operator, $\vf^{l}_{\cdot, 0}$ the approximation subband and $(\vf^{l}_{\cdot, k})_{k=1}^{K_{l}}$ the detail subbands. Here we assume that $\vf^{0}_{c, 0}=\vi_{c}$ and $\vf^{0}_{s, 0}=\vi_{s}$. Note that this formulation encompasses both Wavelet and Contourlet representations.

At each scale, the approximation subband $\vf^{l}_{\cdot, 0}$ captures the overall structure or low-frequency components, and the directional subbands $(\vf^{l}_{\cdot, k})_{k=1}^{K_{l}}$ encode fine-grained details, such as edges and other high-frequency information.

\begin{figure}[t]
\subfloat[Enforcing style details via ETF-based subband fusion $\oplus$ at scale $l$.]{
\begin{minipage}[t]{0.5\columnwidth}
\noindent\fcolorbox{white}{white}{\begin{minipage}{0.45\columnwidth}
\centering\textbf{\scriptsize{Style $\vi_{s}$}}\end{minipage}}\noindent\fcolorbox{white}{white}{\begin{minipage}{0.45\columnwidth}
\centering\textbf{\scriptsize{ETF $\ve_{s}$}}\end{minipage}}

\includegraphics[width=\columnwidth]{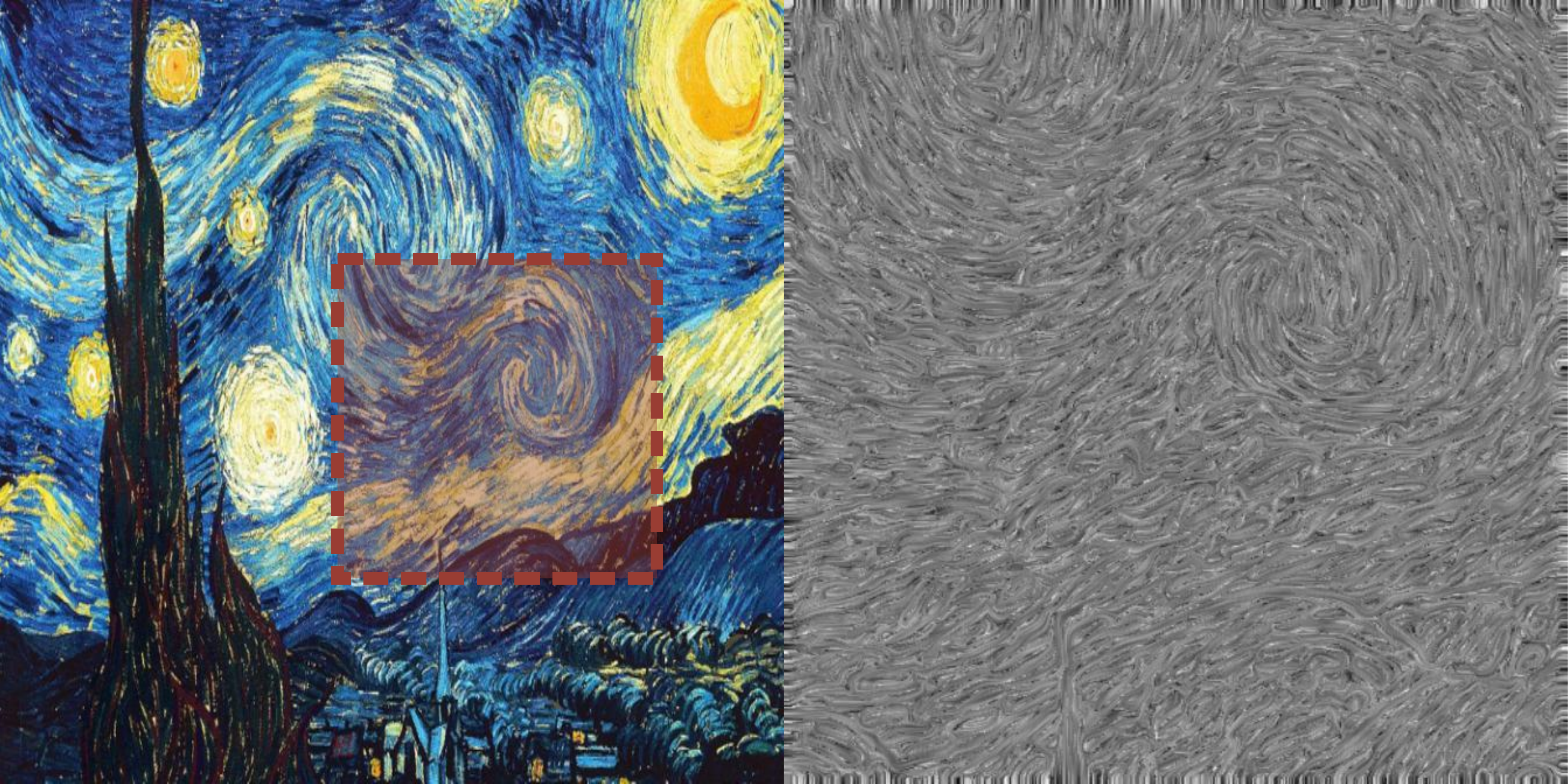}
\end{minipage}\hspace{0.02\columnwidth}
\begin{minipage}[t]{0.473\columnwidth}
\vspace{-0.35 cm}
\includegraphics[width=\columnwidth]{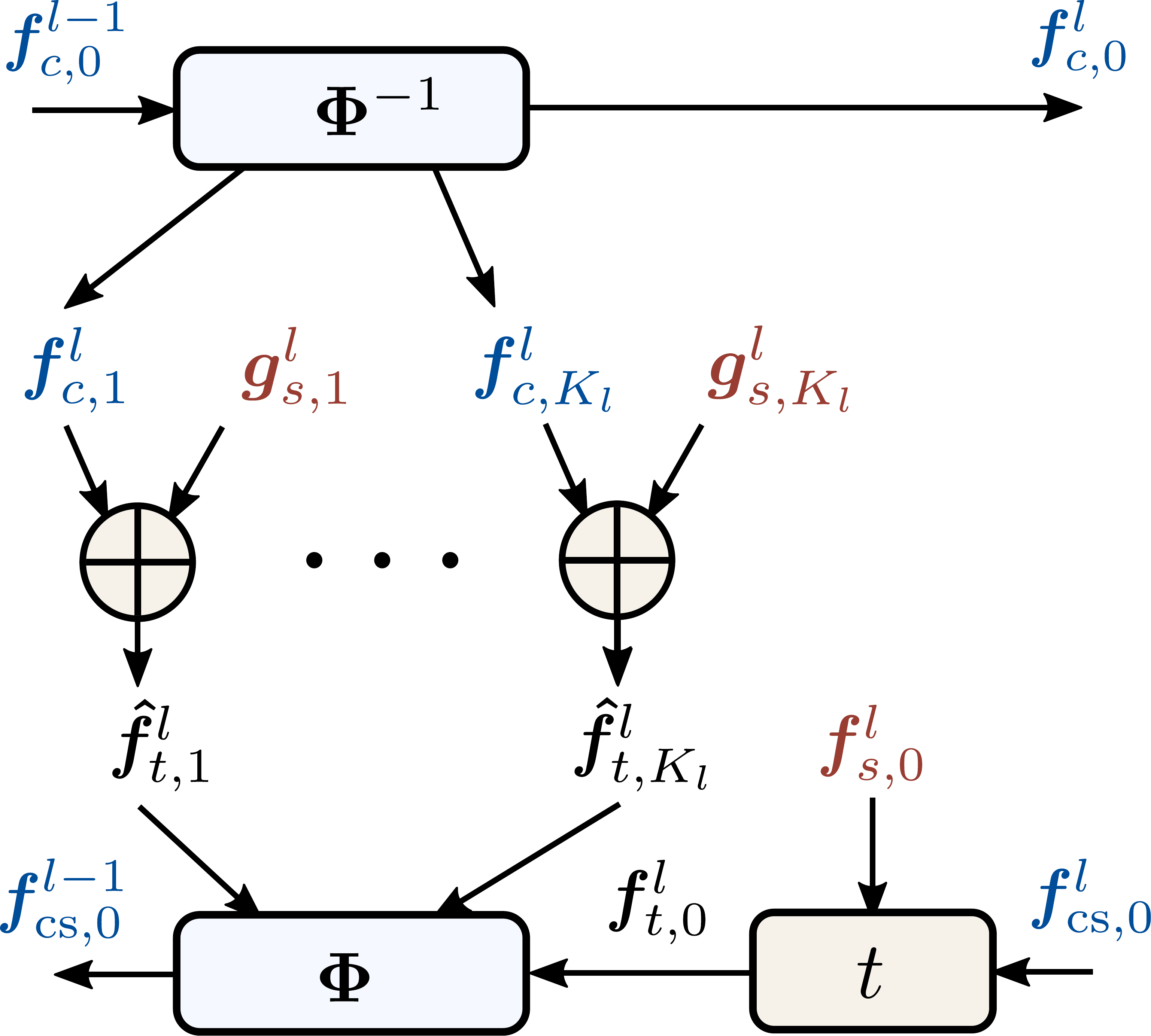}
\end{minipage}}
\vspace{0.325 cm}

\noindent\fcolorbox{white}{white}{\begin{minipage}{0.27\columnwidth}
\centering\textbf{\scriptsize{Content}}
\end{minipage}}\noindent\fcolorbox{white}{white}{\begin{minipage}{0.26\columnwidth}
\centering \textbf{\scriptsize{Photorealistic}}
\end{minipage}}\noindent\fcolorbox{white}{white}{\begin{minipage}{0.26\columnwidth}
\centering \textbf{\scriptsize{Artistic}}
\end{minipage}}\noindent\fcolorbox{white}{white}{\begin{minipage}{0.12\columnwidth}
\centering \textbf{\scriptsize{\textbf{Zoom-in}}}
\end{minipage}}

\vspace{-0.35 cm}
\subfloat[Effect of ETF-based detail subband fusion.]{
\includegraphics[width=\columnwidth]{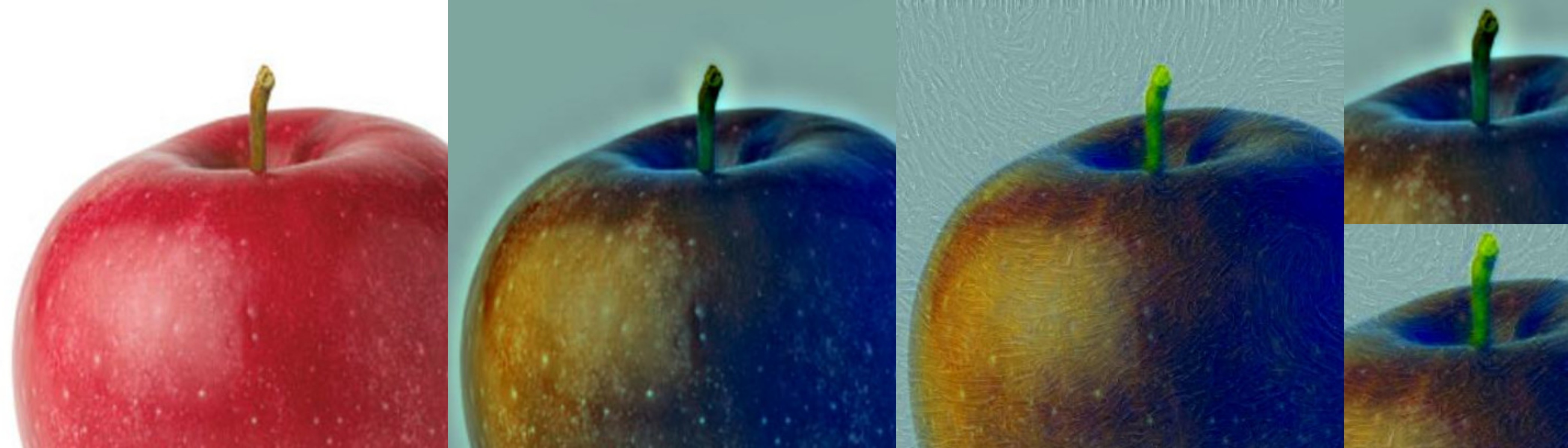}}
\caption{\textbf{Enforcing artistic Style Transfer with subband fusion.} We compute the Edge Tangent Flow of the style image $\vi_{s}$ to extract its detail subbands
$\{\vg^{l}_{s,k>0}\}_{l=1}^{L}$. These are then fused with the corresponding content subbands from coarse to fine scale, promoting an artistic image appearance.}
\label{fig:artistic_diag}
\vspace{-0.17 cm}
\end{figure}

\noindent{\textbf{Subband Matching.}
We align content and style multiresolution subbands using the Wasserstein-$2$ distance under a Gaussian relaxation. At any scale, given content and style representations $(\vf_{c}, \vf_{s})$ with $K$ directions, the aligned representation corresponds to
\begin{align}
  \label{eq:align_general}
  \vf_{t}=&\ \gT(\vf_{c}, \vf_{s})= \begin{pmatrix}
      \vf_{t,0} & \dots & \vf_{t,K}
                  \end{pmatrix}
\end{align}
where the $k$-th aligned subband $\vf_{t,k} \in \mathbb{R}^{C \times W_{c,k} \times H_{c,k}}$ is obtained by matching content coefficients $\vf_{c, k} \in \mathbb{R}^{C \times W_{c,k} \times H_{c,k}}$ with style coefficients $\vf_{s, k} \in \mathbb{R}^{C \times W_{s,k} \times H_{s,k}}$
\begin{align}
 \vf_{t, k}=&\ t(\vf_{c, k}, \vf_{s, k})=\ \text{vec}^{-1} \big(\mT^{\star}_{k}\text{vec}(\vf_{c,k})\big)
\end{align}
Here, $\mT^{\star}_{k}$ corresponds to the optimal transport map associated to the $k$-th content and style subbands. Note that $\mT^{\star}_{k}$ is applied over content datapoints $\text{vec}(\vf_{c,k}) \in \mathbb{R}^{C \times W_{c,k}H_{c,k}}$, where $\text{vec}(\cdot)$ denotes vectorization. This aligns with the Neural Style Transfer paradigm, where each spatial location of a feature map is treated as an individual data point.

Alternatively, recent work has explored the use of optimal transport to compute distances between Gaussian Mixture Models (GMMs) \cite{delon_2020_wasserstein}. This suggests a potential generalization where Wavelet distributions are approximated by GMMs, aligning with prior work on Wavelet subband modeling \cite{do_2002_rotation}. We defer this direction to future research.

\begin{figure}[t]
\vspace{-0.05 cm}
\centering
\subfloat[Combination of style at scale $l$ using a single style reference $\vi_{s}^{1}$.]{
\includegraphics[width=0.925\columnwidth]{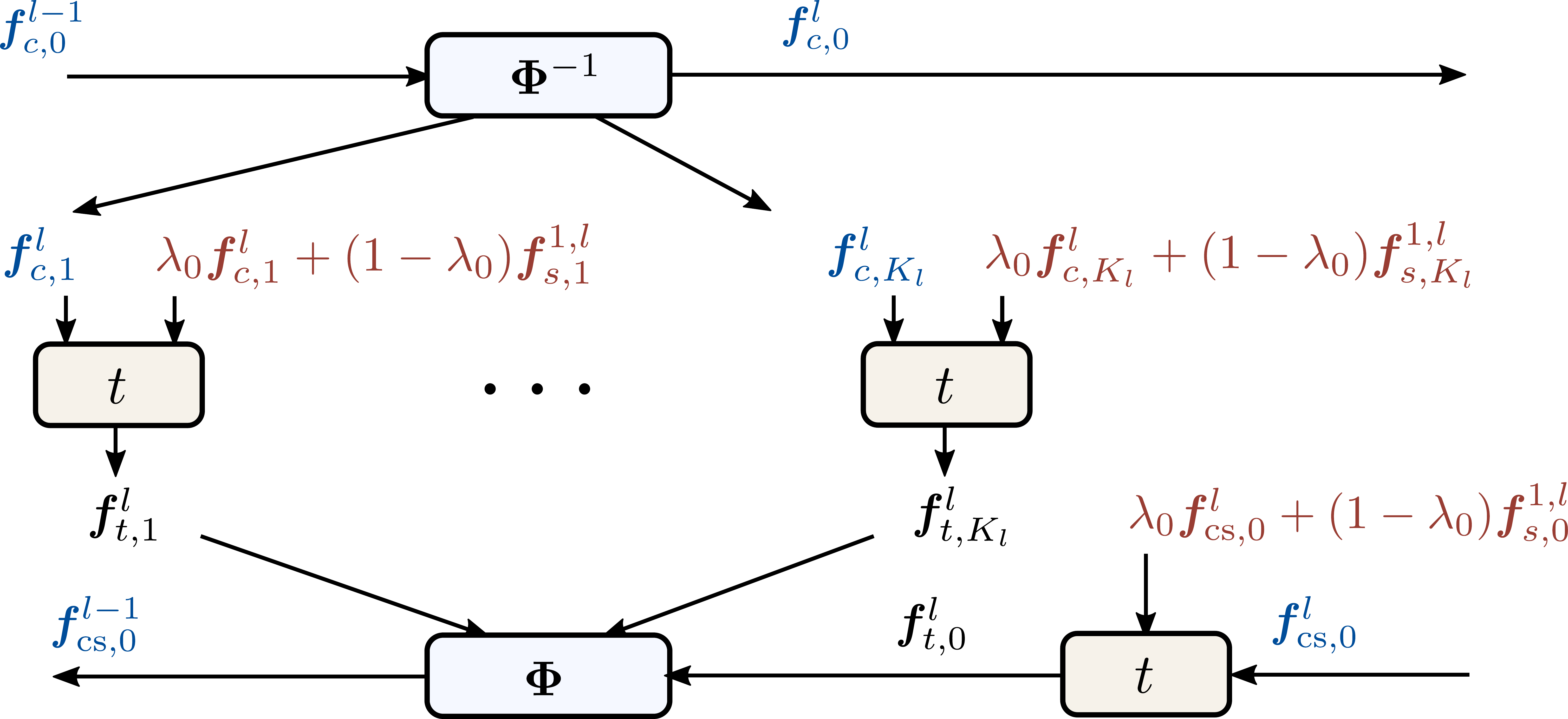}}
\vspace{0.275 cm}

\noindent\fcolorbox{white}{white}{\begin{minipage}{0.23\columnwidth}
\centering\textbf{\scriptsize{Content $\vi_{c}$}}
\end{minipage}}\noindent\fcolorbox{white}{white}{\begin{minipage}{0.23\columnwidth}
\centering \textbf{\scriptsize{$\lambda_{0}=0.75$}}
\end{minipage}}\noindent\fcolorbox{white}{white}{\begin{minipage}{0.23\columnwidth}
\centering \textbf{\scriptsize{$\lambda_{0}=0.25$}}
\end{minipage}}\noindent\fcolorbox{white}{white}{\begin{minipage}{0.23\columnwidth}
\centering \textbf{\scriptsize{\textbf{Style $\vi_{s}^{1}$}}}
\end{minipage}}

\vspace{-0.4 cm}
\subfloat[Example of style interpolation using a single style reference $\vi_{s}^{1}$.]{
\includegraphics[width=\columnwidth]{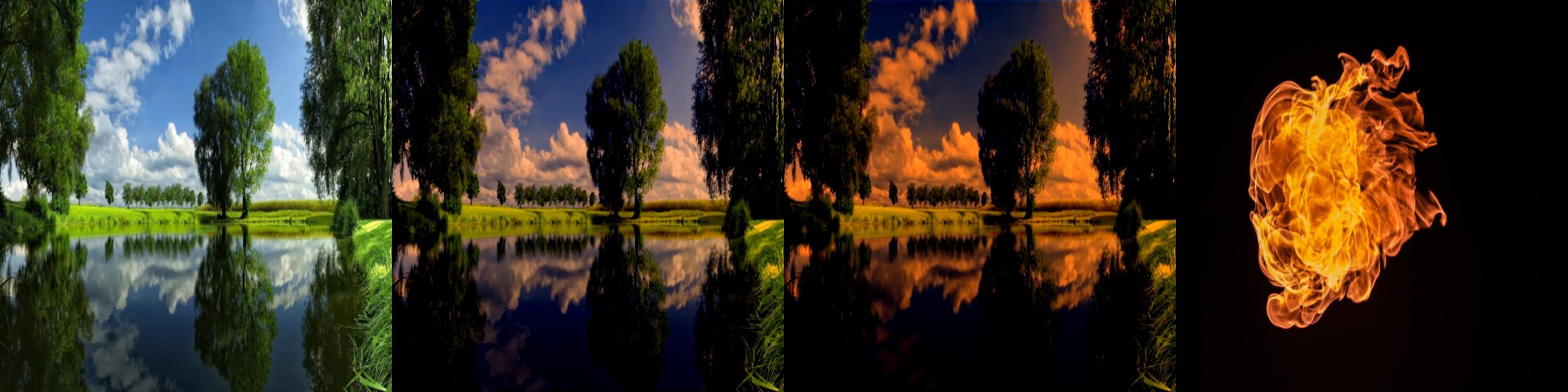}}
\caption{\textbf{Interpolating style in the representation space.} A convex combination of the content and style subbands enables a fine control over the stylization strength. For instance, given a single style reference $\vi^{1}_{s}$ and blending factors $\bm{\lambda}=(\lambda_{0}\ \ 1-\lambda_{0})$, increasing the weight of the content reference $\lambda_{0}$ attenuates the Style Transfer effect.}
\label{fig:interpolation_diag}
\vspace{-0.25cm}
\end{figure}

\noindent{\textbf{Multiresolution Decoder.}
Given multiscale content and style representations $(\vf_{c}, \vf_{s})_{l=1}^{L}$, we progressively align and combine content and style subbands from coarse to fine resolution. This multiscale approach, similar to the \textit{encode-align-decode} strategy in traditional Neural Style Transfer, enables us to match style at multiple levels of detail. However, by leveraging Wavelet and Contourlet operators, we avoid the need for generating intermediate images or training image decoders, directly synthesizing stylized representations in a single step.

At the coarsest level $L$, let the \textit{intermediate} stylized representation $\vf^{L}_{\text{cs}}$ be defined as the content representation itself:
\begin{align}
    \label{eq:decoder_initialization}
    \vf^{L}_{\text{cs}}=&\ \vf^{L}_{c}=\ \begin{pmatrix}
      \vf^{L}_{c,0} & \dots & \vf^{L}_{c,K_{L}}
                  \end{pmatrix}
\end{align}

Next, following Eq. \eqref{eq:align_general}, we align the stylized $\vf^{L}_{\text{cs}}$ and style $\vf^{L}_{s}$ representations to obtain an aligned representation
\begin{align}
    \vf^{L}_{t}=&\ \gT(\vf^{L}_{\text{cs}}, \vf^{L}_{s})
\end{align}
with subbands corresponding to $\big(\vf^{L}_{t,k}= t(\vf^{L}_{\text{cs},k}, \vf^{L}_{s,k})\big)_{k=0}^{K_{L}}$. Once subbands are aligned, we invert them to obtain an intermediate stylized approximation subband at scale $L-1$
\begin{align}
  \vf^{L-1}_{\text{cs},0}=&\ \bm{\Phi}\vf^{L}_{t}
\end{align}

We repeat the process at every scale $l \in \{L-1, \dots, 1\}$ from coarse to fine resolution, where the stylized representation $\vf^{l}_{\text{cs}}$ consists of the stylized approximation and the content details
\begin{align}
    \label{eq:stylized_representation}
    \vf^{l}_{\text{cs}}=& \begin{pmatrix}
      \vf^{l}_{\text{cs},0} & \vf^{l}_{c,1} & \dots & \vf^{l}_{c,K_{l}}
                  \end{pmatrix}
\end{align}
The aligned representation $\vf^{l}_{t}$ is then obtained by matching stylized $\vf^{l}_{\text{cs}}$ and style representations $\vf^{l}_{s}$
\begin{align}
  \label{eq:representation_alignment}
  \vf^{l}_{t}=& \gT(\vf^{l}_{\text{cs}},\vf^{l}_{s})
\end{align}
and the finer-scale approximation is computed via synthesis
\begin{align}
  \label{eq:finer_approximation}
  \vf^{l-1}_{\text{cs},0}=&\ \bm{\Phi}\vf^{l}_{t}
\end{align}
Finally, the stylized image is obtained by inverting the finest aligned representation
\begin{align}
  \label{eq:stylized_image}
  \vi_{\text{cs}}=&\ \bm{\Phi}\vf^{1}_{t} \in \mathbb{R}^{C \times W_{c}\times H_{c}}
\end{align}
\algref{alg:gist} presents the detailed steps of our Style Transfer approach, which leverages multiscale geometric image representations. \figref{fig:gist_diag} provides a visual illustration of the proposed multiscale representation alignment process.

\subsection{Artistic Style Transfer via Edge Tangent Flow}
\label{sec:artistic_style_transfer}
We extend GIST to artistic Style Transfer by amplifying the textural details of the style image and integrating them with the content details in the representation space. Specifically, we extract the Edge Tangent Flow (ETF) \cite{kang_2008_flow} of the style image, compute its multiscale representation, and align the content detail subbands with it using a fusion-based strategy.

Edge Tangent Flow is a non-photorealistic rendering technique that efficiently produces a smooth, stylized edge vector field. It operates by smoothing an image's gradient, emphasizing salient edge directions, and locally aligning weaker edges with dominant ones. Given an image $\vi \in \mathbb{R}^{W \times H}$, its ETF $\ve \in \mathbb{R}^{W \times H}$ can be iteratively defined as:
\begin{align}
  \ve^{(j+1)}[\vx]=&\frac{1}{\tau}\sum_{\vy \in \Omega_{\vx}}\phi_{\vx, \vy}\ve^{(j)}[\vy]\ w^{s}_{\vx, \vy}w^{m}_{\vx, \vy}w^{d}_{\vx, \vy}
\end{align}
where $\Omega_{\vx}$ denotes the neighborhood of coordinate $\vx \in \mathbb{Z}^{2}$, $\tau$ a normalization factor, $\phi_{\vx, \vy}\in \{-1, 1\}$ a vector alignment function, $w^{s}_{\vx, \vy}$ a spatial weight function, $w^{m}_{\vx, \vy}$ a magnitude weight function, and $w^{d}_{\vx, \vy}$ a direction weight function. For a comprehensive explanation, refer to \citet{kang_2007_coherent, kang_2008_flow}.

Let $\ve_{s}$ denote the ETF of the style image $\vi_{s}$, and $(\vg^{l}_{s})_{l=1}^{L}$ its multiscale representation. To incorporate the shape information captured by $\ve_{s}$ into the content image, we replace the original optimal transport-based alignment of detail subbands with an ETF-based subband fusion technique
\begin{align}
  \bm{\hat{f}}^{l}_{t,k}=&\ \vf^{l}_{\text{cs},k} \oplus \vg^{l}_{s,k},\ k\in\{1,\dots,K_{l}\}
\end{align}
where $\oplus$ denotes the element-wise maximum operation between subbands. Note that the approximation subbands, which represent the global structure, are still aligned using optimal transport. Since content and ETF detail subbands must be the same size, in practice we randomly crop a style image patch, resize it to match the content image dimensions, and then compute its ETF. This allows for diverse stylistic shapes while maintaining consistent spatial and channel dimensions.

By emphasizing the detailed components of the style image through ETF and fusing them with the content details across multiple scales, we can enhance the target texture, promoting an artistic appearance. \figref{fig:artistic_diag} illustrates our proposed artistic Style Transfer method based on ETF subband fusion.

\begin{table}[t]
\centering
\setlength\tabcolsep{4pt}
\begin{center}
\def\arraystretch{1.5}

\caption{\textbf{Photorealistic style transfer performance.} Comparison between photorrealistic Neural Style Transfer techniques and our Geometric-based approach in terms of content preservation, style alignment and inference time.}
\vspace{-1.5\baselineskip}
\label{tab:st_photo_contrast}
\resizebox{\columnwidth}{!}{
\begin{tabular}{c|c|c|c|c|c}
\specialrule{.15em}{.05em}{.05em} 
\makecell{Method} & \makecell{SSIM$\uparrow$\\$(\bm{i}_{\text{cs}}, \bm{i}_{c})$} & \makecell{LPIPS$\downarrow$\\$(\bm{i}_{\text{cs}}, \bm{i}_{s})$} & \makecell{FID$\downarrow$\\$(\bm{i}_{\text{cs}}, \bm{i}_{s})$} & \makecell{Generator\\Trainable Pars.} & \makecell{Inference\\Time (s)}\\
\hline
WCT$^{2}$ (\texttt{Sum}) & $0.6335$ & $\mathbf{0.7593}$ & $199.34$ & $3,505,219$  & $0.37$ \\
WCT$^{2}$ (\texttt{Concat.}) & $0.728$ & $0.7657$ & $194.93$ & $6,601,795$ & $0.45$\\  %
\rowcolor{bg_blue}GIST Wavelets \textbf{(Ours)} & $\mathbf{0.7404}$ & $0.7664$ & $192.43$ & \textbf{None} & $\mathbf{0.1}$\\  %
\rowcolor{bg_blue}GIST Contourlets \textbf{(Ours)} & $0.7323$ & $0.7676$ & $\mathbf{190.74}$ & \textbf{None} & $0.18$\\  %
\specialrule{.15em}{.05em}{.05em} 
\end{tabular}}
\end{center}
\end{table}

\begin{figure}[t]

\noindent\fcolorbox{white}{white}{\begin{minipage}[t]{0.144\columnwidth}
\centering\textbf{\scriptsize{Refs.}}
\end{minipage}}\noindent\fcolorbox{white}{white}{\begin{minipage}[t]{0.144\columnwidth}
\centering \textbf{\scriptsize{Labels}}
\end{minipage}}\noindent\fcolorbox{white}{white}{\begin{minipage}[t]{0.3\columnwidth}
\centering \textbf{\scriptsize{WCT$^{2}$ (\texttt{Concat})}}
\end{minipage}}\noindent\fcolorbox{white}{white}{\begin{minipage}[t]{0.3\columnwidth}
\centering \textbf{\scriptsize{\textcolor{mybb}{\textbf{GIST (Ours)}}}}
\end{minipage}}

\vspace{-0.1 cm}
\includegraphics[width=\columnwidth]{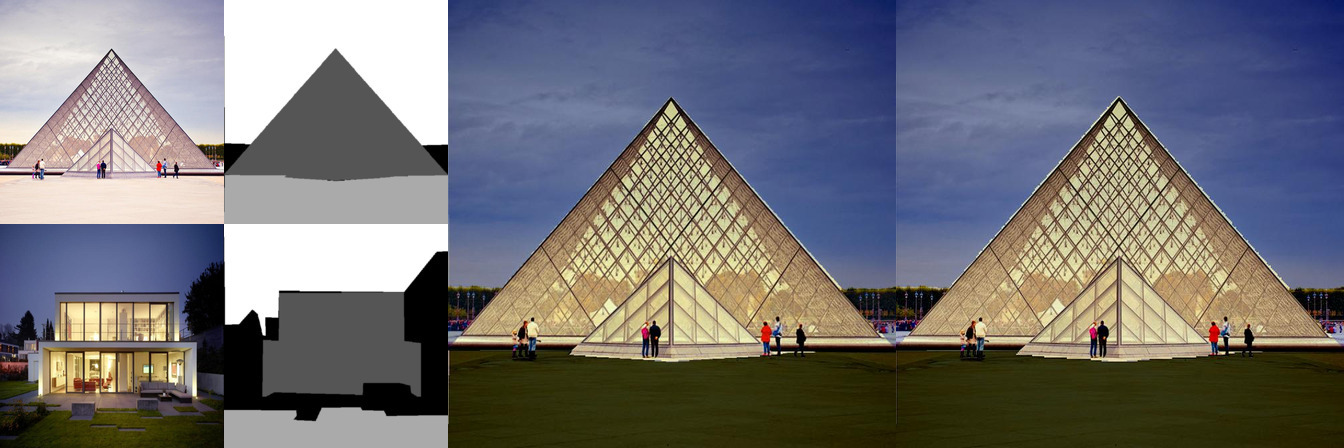}

\includegraphics[width=\columnwidth]{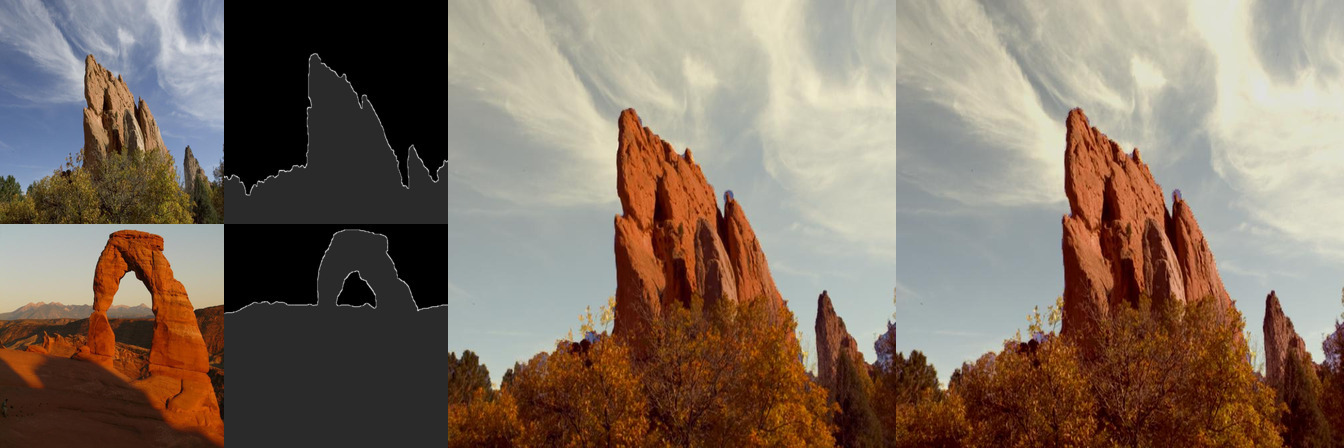}

\includegraphics[width=\columnwidth]{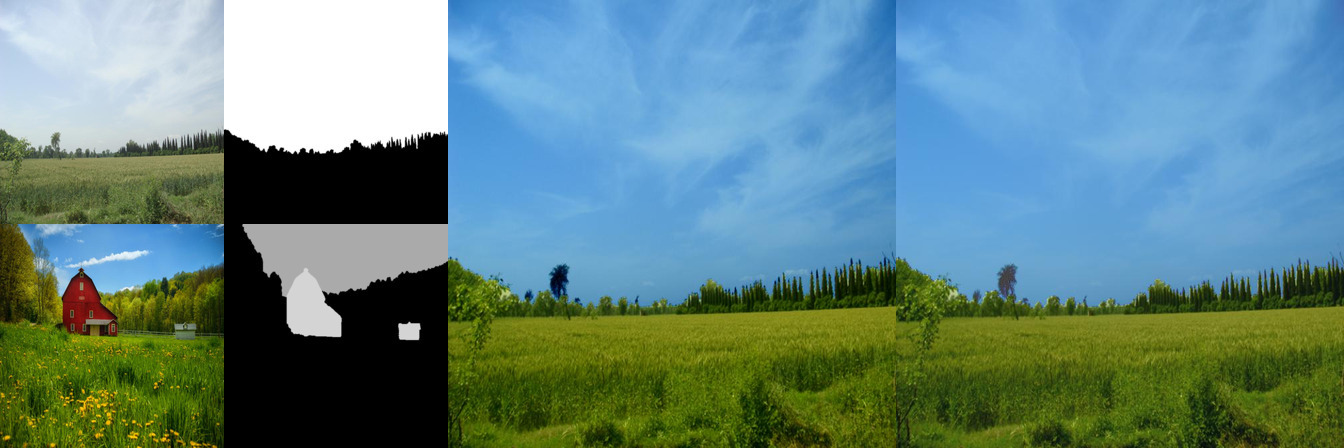}

\vspace{-0.5\baselineskip}
\caption{\textbf{Fine-grained Style Transfer via semantic labels.} Our geometric representation approach allows for targeted stylization using semantic labels, providing control over the Style Transfer process at specific image regions at a fraction of the cost of deep learning methods without sacrificing photorealism.}
\label{fig:st_photo_labels_contrast}
\vspace{-0.01 cm}
\end{figure}

\subsection{Style Interpolation}
\label{sec:interpolation}
Analogous to Neural Style Transfer, decoupling semantic and perceptual attributes across scales enables fine-grained control over the style transferred from a reference image. In particular, blending multiple styles can be achieved by combining style representations prior to the subband alignment.

Given a set of style images $\mS= \big\{\vi^{r}_{s}\big\}_{r=1}^{|\mS|}$ and blending factors $\bm{\lambda}\in [0,1]^{|\mS|+1}$
\begin{align}
  \bm{\lambda}=&\ \begin{pmatrix}\lambda_{0} & \dots & \lambda_{|\mS|}\end{pmatrix},\ \sum\nolimits_{r=0}^{|\mS|}\lambda_{r}=1
\end{align}
a convex combination of styles at scale $l$ can be expressed as
\begin{align*}
    \vf^{l}_{t,k}=&\ t\big(\vf^{l}_{\text{cs},k}, \lambda_{0} \vf^{l}_{\text{cs},k} + \sum\nolimits_{r=1}^{|S|} \lambda_{r} \vf^{r,l}_{s,k}\big)
\end{align*}
where $\vf_{s}^{r,l}$ corresponds to the representation of $\vi_{s}^{r}$ at scale $l$. Note that the style combination, performed prior to the alignment, includes $\vf^{l}_{\text{cs},k}$. This ensures that no stylization occurs for $\lambda_{0}=1$. \figref{fig:interpolation_diag} shows our style interpolation approach for a single style reference.

\section{Experimental Results}
GIST is evaluated using quantitative and qualitative metrics for content and style preservation, as well as processing time. We compare its performance and computational efficiency to deep learning methods and conduct ablation studies to isolate the contributions of individual components and gain deeper insights into our method's effectiveness. All our experiments were conducted using a single NVIDIA Quadro RTX 5000 GPU and PyTorch-only code.

\begin{figure}[t]
\vspace{-0.14 cm}
\noindent\fcolorbox{white}{white}{\begin{minipage}{0.225\columnwidth}
\centering\textbf{\scriptsize{Content}}
\end{minipage}}\noindent\fcolorbox{white}{white}{\begin{minipage}{0.225\columnwidth}
\centering \textbf{\vspace{-0.05 cm}\scriptsize{Style}}
\end{minipage}}\noindent\fcolorbox{white}{white}{\begin{minipage}{0.225\columnwidth}
\centering \textbf{\scriptsize{Photorealistic}}
\end{minipage}}\noindent\fcolorbox{white}{white}{\begin{minipage}{0.225\columnwidth}
\centering \textbf{\scriptsize{\textbf{Artistic}}}
\end{minipage}}

\vspace{-0.05 cm}
\includegraphics[width=\columnwidth]{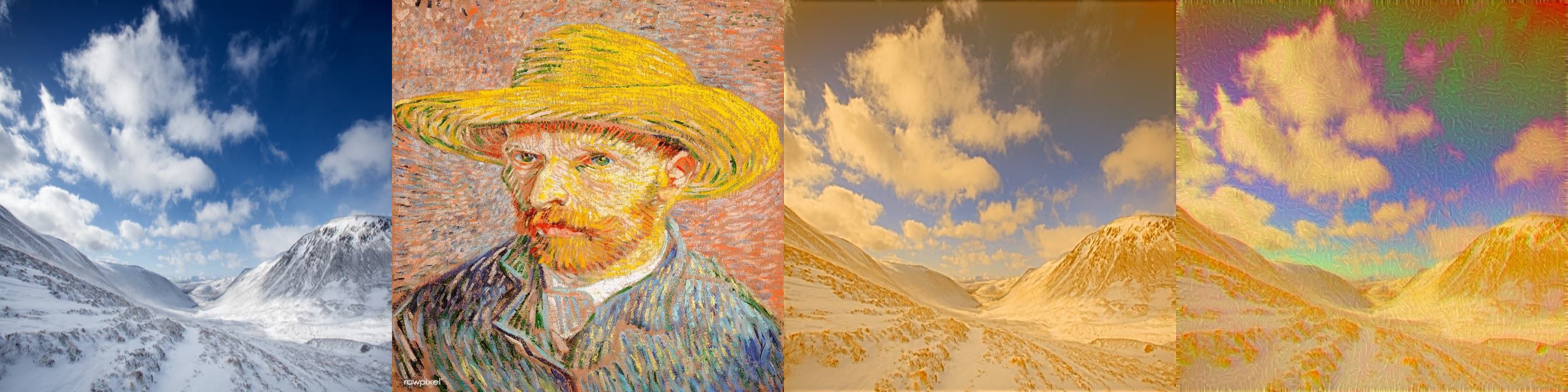}
\includegraphics[width=\columnwidth]{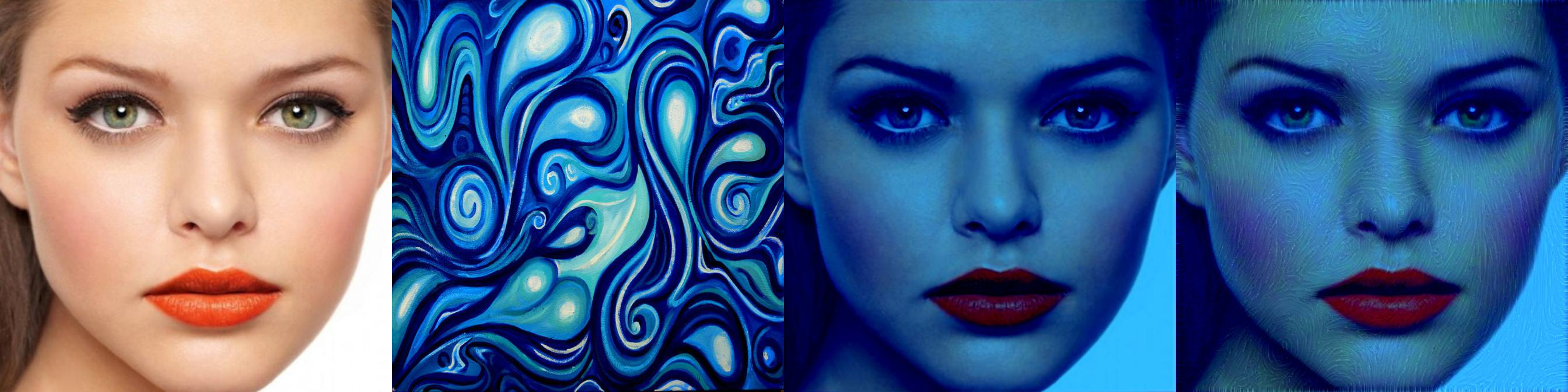}
\includegraphics[width=\columnwidth]{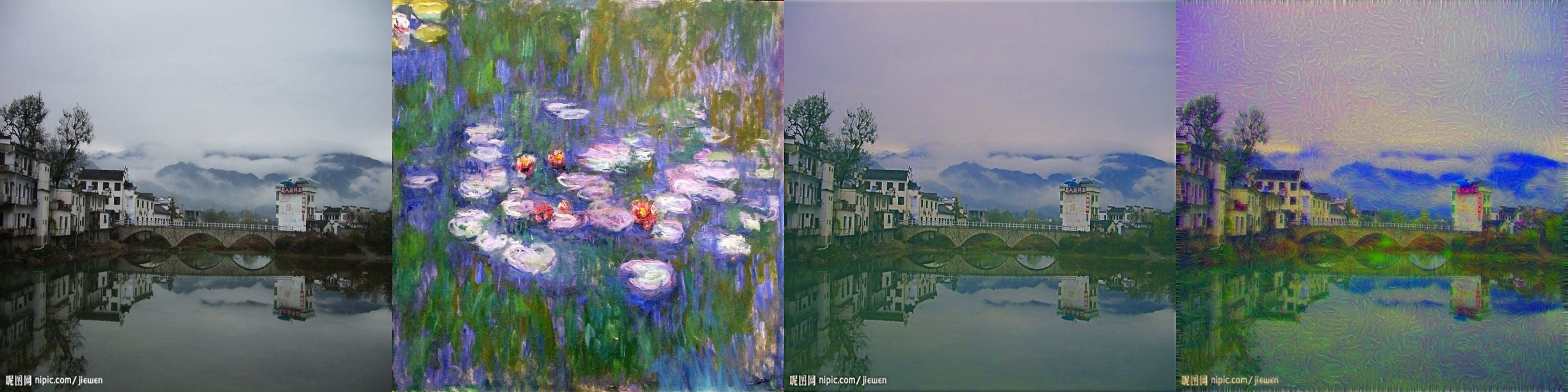}
\includegraphics[width=\columnwidth]{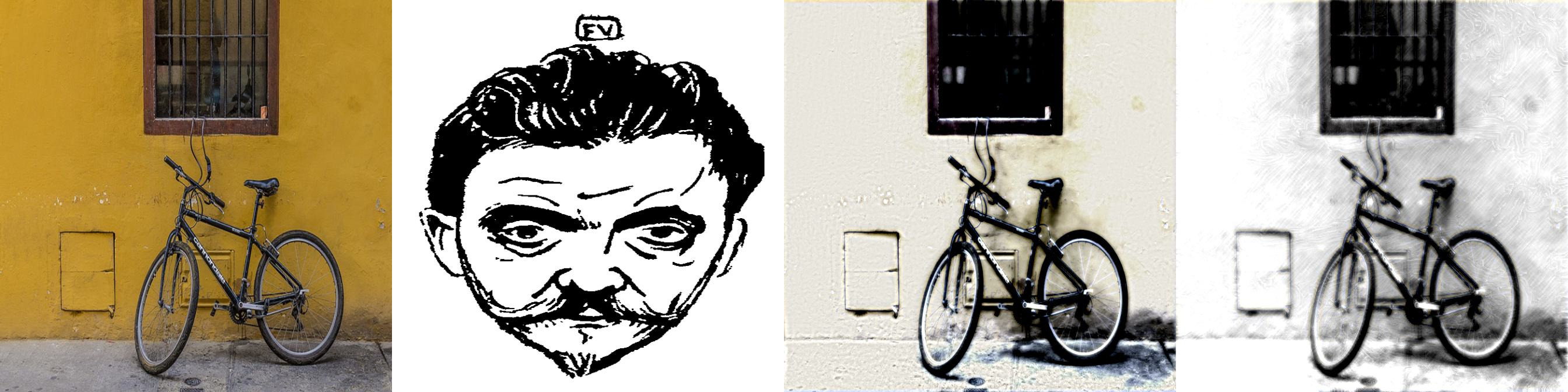}
\vspace{-0.6cm}
\caption{\textbf{Artistic Style Transfer results via geometric representations.} By fusing content and style ETF detail subbands, GIST effectively imposes characteristic shapes of the style onto the stylized output, extending the method's capabilities beyond photorealism to more artistic image transformations.}
\label{fig:artistic_contrast}
\vspace{-0.3cm}
\end{figure}

\subsection{Content and Style Preservation}\label{sec:style_preservation}
As described in Section \ref{sec:proposed_method}, GIST involves extracting content and style geometric coefficients from input images and progressively aligning them across multiple scales. By exploiting the perfect reconstruction property of multiresolution multidirectional representations, we invert the aligned subbands to synthesize photorealistic stylized images. We compare our results to WCT$^2$ \cite{yoo_2019_photorealistic}, a state-of-the-art CNN-based method, in terms of stylization quality and runtime.

\begin{figure}[t]
\includegraphics[width=\columnwidth]{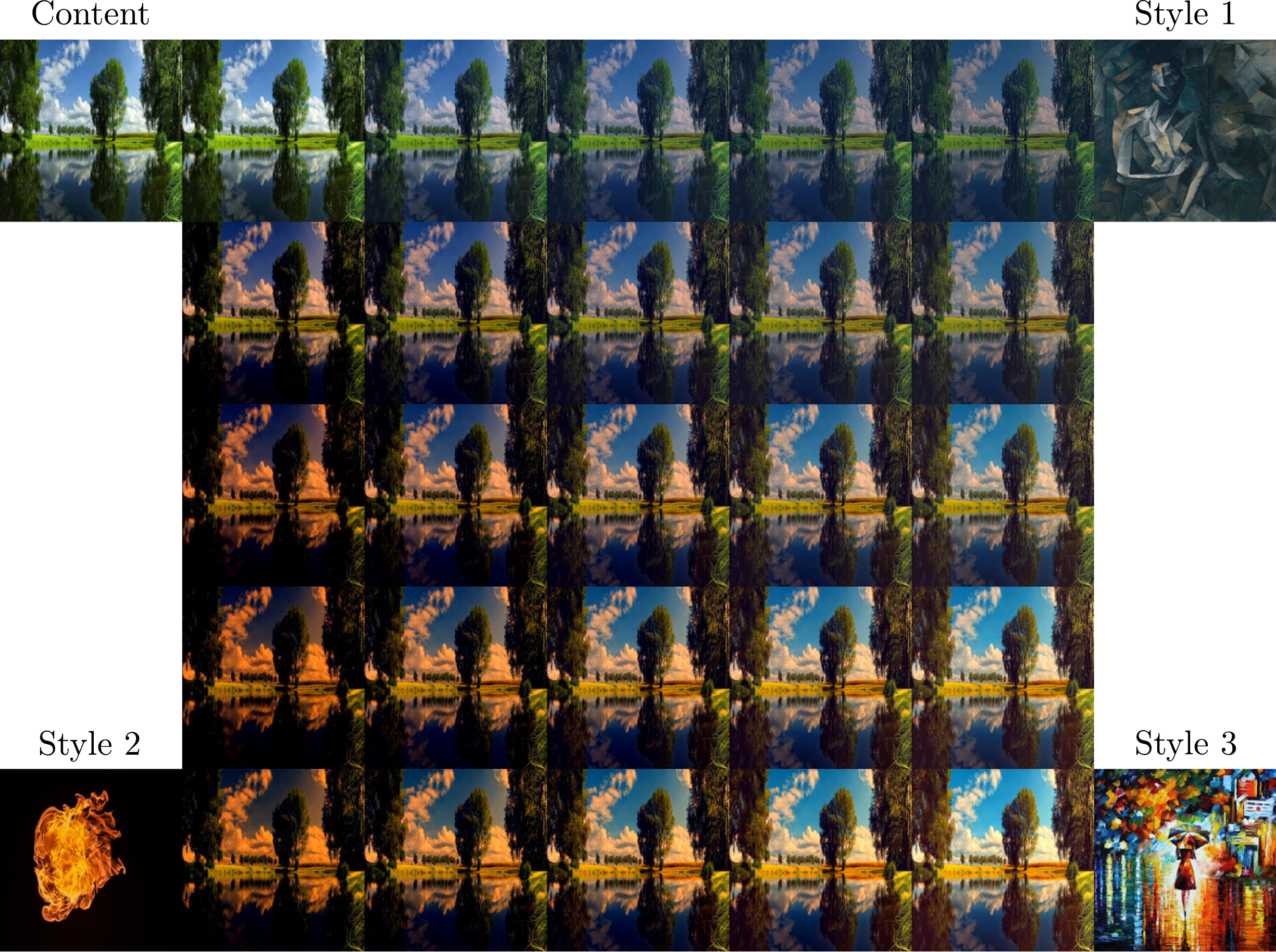}
\vspace{-0.6cm}
\caption{\textbf{Style interpolation results.} Our approach allows obtaining a convex combination of multiple styles in an efficient manner while preserving a natural image appearance by controlling the weight of each style reference in the image representation space.}
\vspace{-0.25 cm}
\label{fig:photo_interpolation}
\end{figure}

{\bf \noindent Setup.} We perform Style Transfer using both Wavelets and Contourlets. For the case of Wavelets, multiresolution representations are obtained using a three scale analysis $(L=3)$ based on the \textit{Daubechies}$-2$ Wavelet. Approximation and detail subbands are extracted based on the Stationary Wavelet Transform (SWT) implementation and aligned independently from each other via optimal transport. For the case of Contourlets, directional subbands are also extracted from $L=3$ scales with configuration $\mK=(1, 4, 4)$ using \textit{pkva} filters. We conduct a comprehensive evaluation of stylized results on a large dataset of $7,500$ image pairs by combining $75$ diverse content images with $100$ distinct style images. All images are RGB rescaled to size $672\times 672$ pixels.

Our method is compared to the two versions of WCT$^{2}$, Wavelet pooling via channel summation (\texttt{Sum}) and concatenation (\texttt{Concat}). We measure the structural similarity (SSIM) \cite{wang_2004_image} between stylized $\bm{i}_{\text{cs}}$ and content $\bm{i}_{c}$ images. Similarly, texture preservation is measured via the Learned Perceptual Image Patch Similarity (LPIPS) \cite{zhang_2018_unreasonable} and Fréchet-Inception Distance (FID) \cite{heusel_2017_gans} between stylized $\bm{i}_{\text{cs}}$ and style $\bm{i}_{s}$ images. Processing time and number of trainable parameters are also reported to measure the computational cost of each method.

{\bf \noindent Results.} \tabref{tab:st_photo_contrast} shows the performance and computational budget of GIST. We obtain on-par results to WCT$^{2}$'s \texttt{Concat}, its best performing version in terms of photorealism. While WCT$^{2}$ obtains slightly higher SSIM, GIST improves in terms of LPIPS and FID. On the other hand, WCT$^{2}$'s \texttt{Sum} obtains marginally better LPIPS at the cost of degrading object shapes.

GIST significantly reduces the Style Transfer's computational cost, requiring less than a third of the inference time needed by WCT$^{2}$'s \texttt{Sum}, its fastest version. Moreover, while WCT$^{2}$ requires training an image generator with at least $3.5$ million trainable parameters, our multiscale representation approach does not require any training. \figref{fig:st_photo_contrast} illustrates the stylization results obtained by our proposed method. GIST achieves comparable stylization results to cutting-edge deep learning methods at a fraction of time.

GIST can also incorporate semantic labels to generate an object-aware Style Transfer, exclusively matching the style between regions of the same category. \figref{fig:st_photo_labels_contrast} shows examples of Style Transfer using segmentation masks. This enables a fine-grained stylization similar to that obtained via deep learning techniques, while reducing its computational budget.

\subsection{Artistic Style Transfer}
To assess the impact of fusing detail subbands with style ETF subbands for artistic Style Transfer purposes, we compare the resulting stylized images to those obtained using the original optimal transport approach designed for photorealism.

{\bf \noindent Setup.} We equip GIST with a Stationary Wavelet Transform using $L=4$ scales and a Daubechies-2 Wavelet basis to obtain multiscale representations. Approximation subbands are aligned using optimal transport, while detail subbands are generated by fusing features extracted from the Edge Tangent Flow (ETF) of the style image, as explained in \secref{sec:artistic_style_transfer}. We found that a few iterations of the edge flow computation process are enough to obtain a stable result. Since ETF is a grayscale image, we use the same ETF across subband channels during the fusion process.

{\bf \noindent Results.} \figref{fig:artistic_contrast} illustrates the results of Style Transfer using ETF-based subband fusion for various images, comparing them to those obtained using our original alignment criterion based on optimal transport. Visually, the results using ETF fusion incorporate texture from the style image while highlighting its color scheme. This is shown in a variety of image regions, including flat areas (e.g., faces and walls), textured areas (e.g., clouds and water), and fine details (e.g., building windows). Overall, these results suggest that refined transformations in the representation space, such as fusion, can be employed to achieve a wider range of style alignment effects based on geometric representations.

\subsection{Style Interpolation}
We qualitatively evaluate the effect of controlling the final stylization by applying a convex combination of style references in the image representation space. Similar to Neural Style Transfer, we demonstrate that GIST enables style interpolation by combining representations prior to the alignment.

{\bf \noindent Setup.} For evaluation purposes, we conduct style interpolation using three style references, following the procedure outlined in Section \ref{sec:interpolation}. We evaluate $25$ different combination scenarios, varying the weights of each style from $0$ to $1$ in increments of $0.25$. In all cases, multiscale representations are extracted using GIST equipped with a Stationary Wavelet Transform based on $L=3$ scales and a Daubechies-2 basis.

\begin{table}[t]
\centering
\setlength\tabcolsep{4pt}
\begin{center}

\def\arraystretch{1.5}

\caption{\textbf{Effect of number of scales.} Evaluation of the Style Transfer performance in terms of structure preservation, texture transferring and inference time for GIST equipped with SWT and varying number of scales.}
\vspace{-0.65\baselineskip}
\label{tab:ablation_scales}
\resizebox{\columnwidth}{!}{
\begin{tabular}{c|c|c|c|c|c} 
\specialrule{.15em}{.05em}{.05em} 
\makecell{Number of scales $(L)$} & $1$ & $2$ & $3$ & $4$ & $5$\\
\hline
\makecell{SSIM$\uparrow$ $(\bm{i}_{\text{cs}}, \bm{i}_{c})$} & $\mathbf{0.7533}$ & $0.7315$ & $0.7237$ & $0.7154$ & $0.7086$\\
\makecell{LPIPS$\downarrow$ $(\bm{i}_{\text{cs}}, \bm{i}_{s})$} & $0.7748$ & $0.7661$ & $\mathbf{0.7625}$ & $0.7646$ & $0.7727$\\
\makecell{Inference Time (s)} & $\mathbf{0.121}$ & $0.1264$ & $0.1374$ & $0.1457$ & $0.1543$\\
\specialrule{.15em}{.05em}{.05em} 
\end{tabular}}
\end{center}
\vspace{-0.5cm}
\end{table}

{\bf \noindent Results.} \figref{fig:photo_interpolation} illustrates style interpolation using multiple reference images. The resulting stylized images seamlessly blend the combined style representations, as evidenced by their photorealistic appearance and smooth style transitions. By applying a convex combination of styles before alignment via optimal transport, we achieve efficient style blending without incurring additional computational costs for alignment. Importantly, this approach ensures that the objects within the scene remain well-preserved across all style combinations.

\subsection{Ablation Study}
To isolate the effects of GIST's components during stylization, we conduct two analyses: (i) the influence of the number of representation scales and (ii) the impact of using a stationary Wavelet transform (SWT) versus a decimated one (DWT).

\begin{figure}[t]

\noindent\fcolorbox{white}{white}{\begin{minipage}{0.065\columnwidth}
\centering\textbf{\scriptsize{Refs.}}
\end{minipage}}\noindent\fcolorbox{white}{white}{\begin{minipage}{0.155\columnwidth}
\centering \textbf{\scriptsize{$L=1$}}
\end{minipage}}\noindent\fcolorbox{white}{white}{\begin{minipage}{0.155\columnwidth}
\centering \textbf{\scriptsize{$L=2$}}
\end{minipage}}\noindent\fcolorbox{white}{white}{\begin{minipage}{0.155\columnwidth}
\centering \textbf{\scriptsize{$L=3$}}
\end{minipage}}\noindent\fcolorbox{white}{white}{\begin{minipage}{0.155\columnwidth}
\centering \textbf{\scriptsize{$L=4$}}
\end{minipage}}\noindent\fcolorbox{white}{white}{\begin{minipage}{0.155\columnwidth}
\centering \textbf{\scriptsize{$L=5$}}
\end{minipage}}

\includegraphics[width=\columnwidth]{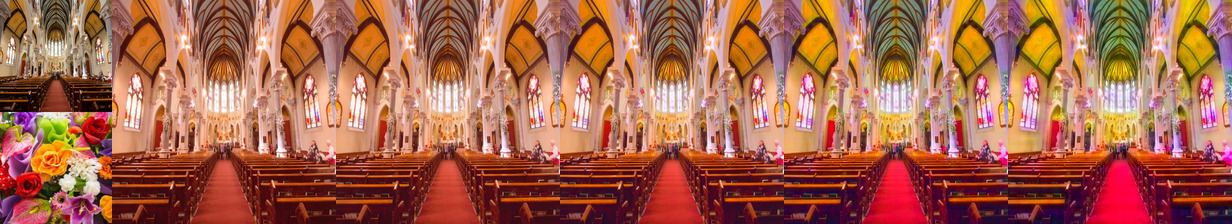}

\includegraphics[width=\columnwidth]{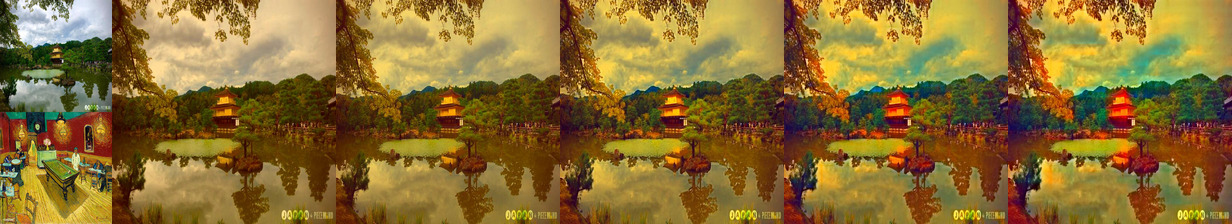}

\includegraphics[width=\columnwidth]{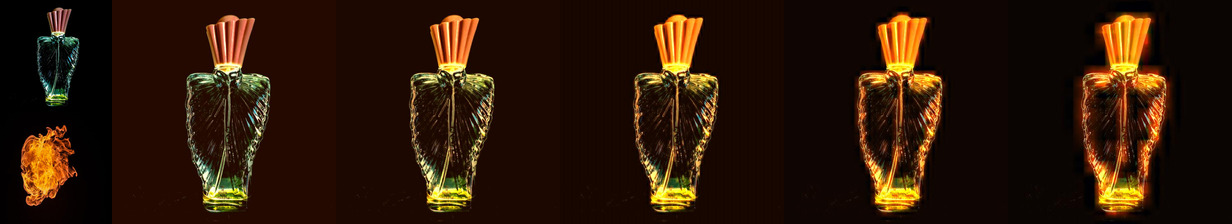}

\vspace{-0.2cm}
\caption{\textbf{Examples of images stylized at different scales.} By varying the number of scales, we can control the trade-off between preserving the content's structure and transferring the style's texture. Increasing scales enhances texture transfer, albeit with a slight increase in computational cost.}
\label{fig:ablation_scales}
\vspace{-0.25cm}
\end{figure}

{\bf \noindent Number of Scales.} GIST introduces the number of representation scales as a hyperparameter. Previous research has shown the significance of multi-scale feature extraction for capturing both fine and coarse texture details \cite{gatys_2017_controlling}. To investigate the influence of decomposition levels on stylization quality and performance, we conduct a comprehensive evaluation.

To measure content and style preservation, we experiment with decomposition levels $L$ ranging from $1$ to $5$ using the Daubechies-2 basis. As in \secref{sec:style_preservation}, we employ SSIM to assess content preservation by comparing the content and stylized images. Similarly, LPIPS is used to quantify the texture preservation by comparing the style and stylized images. We evaluate stylization on the same dataset as in \secref{sec:style_preservation}. In addition to stylization quality, we also compare the computational cost for different number of scales.

As shown in \tabref{tab:ablation_scales}, the number of representation scales significantly impacts Style Transfer results. Increasing the number of scales leads to improved texture preservation compared to the single-scale scenario, while reducing the number of scales helps to maintain the content's structure. Although additional scales slightly increase processing time, they can substantially enhance the overall quality of the stylized image.

\figref{fig:ablation_scales} presents stylization examples for various representation scales. Visually, images stylized with fewer scales closely resemble the content image in both shape and texture, suggesting limited texture extraction for small $L$ values. Conversely, images stylized with more scales capture and impose style attributes more effectively. Both quantitative metrics and visual examples demonstrate a trade-off between content and style preservation as the number of scales increases.

\begin{table}[t]
\centering
\setlength\tabcolsep{4pt}
\begin{center}
\def\arraystretch{1.5}

\caption{\textbf{Effect of stationary Wavelet representations.} Comparison of critically-sampled (DWT) and oversampled (SWT) representations for photorealistic Style Transfer.}
\vspace{-1.5\baselineskip}
\label{tab:ablation_stationary}
\resizebox{\columnwidth}{!}{
\begin{tabular}{c|c|c|c|c|c}
\specialrule{.15em}{.05em}{.05em} 
\makecell{Method} & Undecimated & \makecell{SSIM$\uparrow$\\$(\bm{i}_{\text{cs}}, \bm{i}_{c})$} & \makecell{LPIPS$\downarrow$\\$(\bm{i}_{\text{cs}}, \bm{i}_{s})$} & \makecell{FID$\downarrow$\\$(\bm{i}_{\text{cs}}, \bm{i}_{s})$} & \makecell{Inference\\Time (s)}\\
\hline
DWT $(L=1)$ & \xmark & $0.7473$ & $0.772$ & $196.1$ & $\bm{0.0081}$\\
DWT $(L=4)$ & \xmark & $0.7203$ & $\bm{0.7659}$ & $\bm{190.10}$ & $0.0293$\\
SWT $(L=1)$ & \cmark & $\bm{0.7477}$ & $0.772$ & $196.14$ & $0.036$\\
SWT $(L=4)$ & \cmark & $0.7404$ & $0.7664$ & $192.43$ & $0.1$\\
\specialrule{.15em}{.05em}{.05em} 
\end{tabular}}
\end{center}
\vspace{-0.5cm}
\end{table}

{\bf \noindent Decimated vs. Stationary Wavelets.} GIST can leverage critically sampled representations like DWT and oversampled ones like SWT. We explore how SWT mitigates the artifacts arising from the sampling operations inherent in DWT.

To assess stylization performance, we employ SSIM to measure content preservation, as well as LPIPS and FID to measure style preservation. To investigate the impact of down and upsampling, we compare DWT and SWT representations for both single ($L=1$) and multi-scale ($L=4$) scenarios. We also evaluate the computational cost of each case.

\tabref{tab:ablation_stationary} compares GIST performance using DWT and SWT representations. For a single scale, both achieve similar results in terms of stylization, with DWT offering faster processing time. For multiple scales ($L=4$), aligning decimated representations degrades structural similarity, while undecimated representations better preserve structure and texture, as measured by LPIPS and FID. Despite slightly increasing inference time, matching SWT representations remain significantly faster than deep learning methods such as WCT$^{2}$.

\figref{fig:ablation_stationary} shows GIST Style Transfer results using DWT and SWT multiscale representations. As expected, DWT leads to visible artifacts, while SWT effectively mitigates these distortions due to its non-subsampled implementation. Despite the similar stylization results for DWT and SWT for a single scale, the progressive subsampling in the critically sampled case severely affects the final image appearance.

\section{Conclusion}
We propose GIST, a novel technique for photorealistic Style Transfer based on geometric multiscale image representations. GIST achieves photorealistic stylized images on par or superior to deep learning techniques, without requiring any training or pre-trained models and at a fraction of their computational cost. GIST is a general framework accommodating multiscale and multidirectional representations such as Wavelets and Contourlets, offering fine-grained control over scales and style weights. We extend GIST to artistic Style Transfer by incorporating the style's Edge Tangent Flow to enforce stylistic shapes, demonstrating its versatility. Our experiments quantitatively and qualitatively validate GIST's advantages over alternative deep learning methods.

\begin{figure}[t]

\noindent\fcolorbox{white}{white}{\begin{minipage}{0.14\columnwidth}
\centering\textbf{\vspace{0.05cm}\scriptsize{Content}}
\end{minipage}}\noindent\fcolorbox{white}{white}{\begin{minipage}{0.14\columnwidth}
\centering \textbf{\scriptsize{Style}}
\end{minipage}}\noindent\fcolorbox{white}{white}{\begin{minipage}{0.14\columnwidth}
\centering \textbf{\scriptsize{DWT$_{L=1}$}}
\end{minipage}}\noindent\fcolorbox{white}{white}{\begin{minipage}{0.14\columnwidth}
\centering \textbf{\scriptsize{SWT$_{L=1}$}}
\end{minipage}}\noindent\fcolorbox{white}{white}{\begin{minipage}{0.14\columnwidth}
\centering \textbf{\scriptsize{DWT$_{L=4}$}}
\end{minipage}}\noindent\fcolorbox{white}{white}{\begin{minipage}{0.14\columnwidth}
\centering \textbf{\scriptsize{SWT$_{L=4}$}}
\end{minipage}}

\includegraphics[width=\columnwidth]{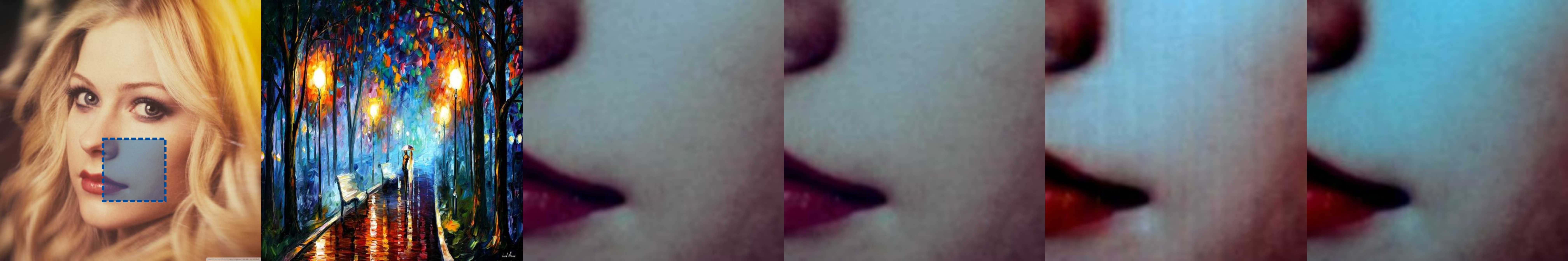}

\includegraphics[width=\columnwidth]{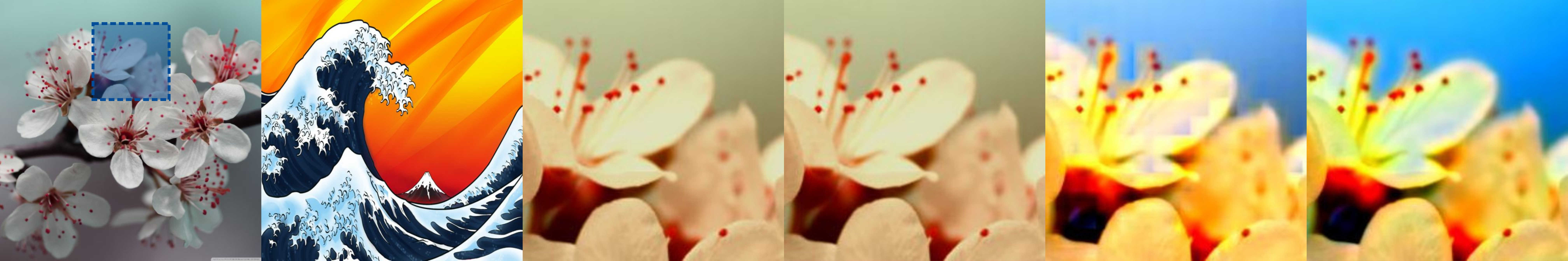}

\includegraphics[width=\columnwidth]{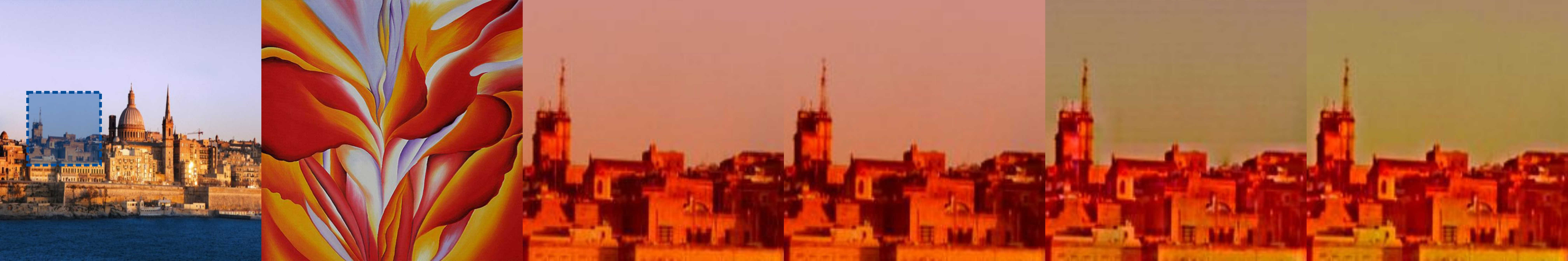}
\vspace{-0.6cm}
\caption{\textbf{Examples of Style Transfer with and without decimation.} Decimated representations extracted via DWT struggle to accurately capture style details with fewer scales and are prone to distortions with more scales (DWT$_{L=4}$). Multiscale undecimated representations obtained via SWT (SWT$_{L=4}$) avoid such distorsions, allowing their use towards photorealism.}
\label{fig:ablation_stationary}
\vspace{-1\baselineskip}
\end{figure}

\ifCLASSOPTIONcaptionsoff
  \newpage
\fi

\bibliographystyle{IEEEtranN}
\bibliography{refs.bib}

\begin{thebibliography}{43}
\providecommand{\natexlab}[1]{#1}
\providecommand{\url}[1]{#1}
\csname url@samestyle\endcsname
\providecommand{\newblock}{\relax}
\providecommand{\bibinfo}[2]{#2}
\providecommand{\BIBentrySTDinterwordspacing}{\spaceskip=0pt\relax}
\providecommand{\BIBentryALTinterwordstretchfactor}{4}
\providecommand{\BIBentryALTinterwordspacing}{\spaceskip=\fontdimen2\font plus
\BIBentryALTinterwordstretchfactor\fontdimen3\font minus
  \fontdimen4\font\relax}
\providecommand{\BIBforeignlanguage}[2]{{%
\expandafter\ifx\csname l@#1\endcsname\relax
\typeout{** WARNING: IEEEtranN.bst: No hyphenation pattern has been}%
\typeout{** loaded for the language `#1'. Using the pattern for}%
\typeout{** the default language instead.}%
\else
\language=\csname l@#1\endcsname
\fi
#2}}
\providecommand{\BIBdecl}{\relax}
\BIBdecl

\bibitem[Zhu et~al.(2000)Zhu, Liu, and Wu]{zhu_2000_exploring}
S.~C. Zhu, X.~W. Liu, and Y.~N. Wu, ``Exploring texture ensembles by efficient
  {M}arkov chain monte carlo-toward a ``trichromacy'' theory of texture,''
  \emph{IEEE Transactions on Pattern Analysis and Machine Intelligence},
  vol.~22, no.~6, pp. 554--569, 2000.

\bibitem[Portilla and Simoncelli(2000)]{portilla_2000_parametric}
J.~Portilla and E.~P. Simoncelli, ``A parametric texture model based on joint
  statistics of complex wavelet coefficients,'' \emph{International Journal of
  Computer Vision}, vol.~40, no.~1, pp. 49--70, 2000.

\bibitem[Heeger and Bergen(1995)]{heeger_1995_pyramid}
D.~J. Heeger and J.~R. Bergen, ``Pyramid-based texture analysis/synthesis,'' in
  \emph{Proceedings of the Conference on Computer Graphics and Interactive
  Techniques}, 1995, pp. 229--238.

\bibitem[Gatys et~al.(2017)Gatys, Ecker, Bethge, Hertzmann, and
  Shechtman]{gatys_2017_controlling}
L.~A. Gatys, A.~S. Ecker, M.~Bethge, A.~Hertzmann, and E.~Shechtman,
  ``Controlling perceptual factors in neural style transfer,'' in
  \emph{Proceedings of the IEEE Conference on Computer Vision and Pattern
  Recognition}, 2017, pp. 3985--3993.

\bibitem[Li et~al.(2017)Li, Fang, Yang, Wang, Lu, and Yang]{li_2017_universal}
Y.~Li, C.~Fang, J.~Yang, Z.~Wang, X.~Lu, and M.-H. Yang, ``Universal style
  transfer via feature transforms,'' in \emph{Proceedings of the 31st
  International Conference on Neural Information Processing Systems}, 2017, pp.
  385--395.

\bibitem[Johnson et~al.(2016)Johnson, Alahi, and
  Fei-Fei]{johnson_2016_perceptual}
J.~Johnson, A.~Alahi, and L.~Fei-Fei, ``Perceptual losses for real-time style
  transfer and super-resolution,'' in \emph{European Conference on Computer
  Vision}, 2016, pp. 694--711.

\bibitem[Luan et~al.(2017)Luan, Paris, Shechtman, and Bala]{luan_2017_deep}
F.~Luan, S.~Paris, E.~Shechtman, and K.~Bala, ``Deep photo style transfer,'' in
  \emph{Proceedings of the IEEE Conference on Computer Vision and Pattern
  Recognition}, 2017, pp. 4990--4998.

\bibitem[Mechrez et~al.(2017)Mechrez, Shechtman, and
  Zelnik-Manor]{mechrez_2017_photorealistic}
R.~Mechrez, E.~Shechtman, and L.~Zelnik-Manor, ``Photorealistic style transfer
  with screened poisson equation,'' \emph{British Machine Vision Conference},
  2017.

\bibitem[Yoo et~al.(2019)Yoo, Uh, Chun, Kang, and Ha]{yoo_2019_photorealistic}
J.~Yoo, Y.~Uh, S.~Chun, B.~Kang, and J.-W. Ha, ``Photorealistic style transfer
  via wavelet transforms,'' in \emph{Proceedings of the IEEE International
  Conference on Computer Vision}, 2019, pp. 9036--9045.

\bibitem[An et~al.(2020)An, Xiong, Huan, and Luo]{an_2020_ultrafast}
J.~An, H.~Xiong, J.~Huan, and J.~Luo, ``Ultrafast photorealistic style transfer
  via neural architecture search,'' in \emph{Proceedings of the AAAI Conference
  on Artificial Intelligence}, vol.~34, 2020, pp. 10\,443--10\,450.

\bibitem[Daubechies(1992)]{daubechies_1992_ten}
I.~Daubechies, \emph{Ten Lectures on Wavelets}.\hskip 1em plus 0.5em minus
  0.4em\relax Society for Industrial and Applied Mathematics, 1992.

\bibitem[Do and Vetterli(2005)]{do_2005_contourlet}
M.~N. Do and M.~Vetterli, ``The contourlet transform: an efficient directional
  multiresolution image representation,'' \emph{IEEE Transactions on image
  processing}, vol.~14, no.~12, pp. 2091--2106, 2005.

\bibitem[Candes et~al.(2006)Candes, Demanet, Donoho, and
  Ying]{candes_2006_fast}
E.~Candes, L.~Demanet, D.~Donoho, and L.~Ying, ``Fast discrete curvelet
  transforms,'' \emph{Multiscale Modeling \& Simulation}, vol.~5, no.~3, pp.
  861--899, 2006.

\bibitem[Efros and Leung(1999)]{efros_1999_texture}
A.~A. Efros and T.~K. Leung, ``Texture synthesis by non-parametric sampling,''
  in \emph{Proceedings of the IEEE International Conference on Computer
  Vision}, vol.~2.\hskip 1em plus 0.5em minus 0.4em\relax IEEE, 1999, pp.
  1033--1038.

\bibitem[Wei and Levoy(2000)]{wei_2000_fast}
L.-Y. Wei and M.~Levoy, ``Fast texture synthesis using tree-structured vector
  quantization,'' in \emph{Proceedings of the Conference on Computer Graphics
  and Interactive Techniques}, 2000, pp. 479--488.

\bibitem[Fan and Xia(2003)]{fan_2003_wavelet}
G.~Fan and X.-G. Xia, ``Wavelet-based texture analysis and synthesis using
  hidden {M}arkov models,'' \emph{IEEE Transactions on Circuits and Systems I:
  Fundamental Theory and Applications}, vol.~50, no.~1, pp. 106--120, 2003.

\bibitem[Do and Vetterli(2002{\natexlab{a}})]{do_2002_wavelet}
M.~N. Do and M.~Vetterli, ``Wavelet-based texture retrieval using generalized
  gaussian density and kullback-leibler distance,'' \emph{IEEE Transactions on
  Image Processing}, vol.~11, no.~2, pp. 146--158, 2002.

\bibitem[Brochard and Zhang(2022)]{brochard_2022_generalized}
A.~Brochard and S.~Zhang, ``Generalized rectifier wavelet covariance models for
  texture synthesis,'' in \emph{International Conference on Learning
  Representations}, 2022.

\bibitem[Heitz et~al.(2021)Heitz, Vanhoey, Chambon, and
  Belcour]{heitz_2021_sliced}
E.~Heitz, K.~Vanhoey, T.~Chambon, and L.~Belcour, ``A sliced {W}asserstein loss
  for neural texture synthesis,'' in \emph{Proceedings of the IEEE/CVF
  Conference on Computer Vision and Pattern Recognition}, 2021, pp. 9412--9420.

\bibitem[Wang et~al.(2020)Wang, Zhao, Chen, Qiu, Mo, Lin, Xing, and
  Lu]{wang_2020_diversified}
Z.~Wang, L.~Zhao, H.~Chen, L.~Qiu, Q.~Mo, S.~Lin, W.~Xing, and D.~Lu,
  ``Diversified arbitrary style transfer via deep feature perturbation,'' in
  \emph{Proceedings of the IEEE/CVF Conference on Computer Vision and Pattern
  Recognition}, 2020, pp. 7789--7798.

\bibitem[Simonyan and Zisserman(2015)]{simonyan_2015_very}
K.~Simonyan and A.~Zisserman, ``Very deep convolutional networks for
  large-scale image recognition,'' in \emph{International Conference on
  Learning Representations}, 2015.

\bibitem[Huang and Belongie(2017)]{huang_2017_arbitrary}
X.~Huang and S.~Belongie, ``Arbitrary style transfer in real-time with adaptive
  instance normalization,'' in \emph{Proceedings of the IEEE International
  Conference on Computer Vision}, 2017, pp. 1501--1510.

\bibitem[Gatys et~al.(2016)Gatys, Ecker, and Bethge]{gatys_2016_image}
L.~A. Gatys, A.~S. Ecker, and M.~Bethge, ``Image style transfer using
  convolutional neural networks,'' in \emph{Proceedings of the IEEE Conference
  on Computer Vision and Pattern Recognition}, 2016, pp. 2414--2423.

\bibitem[Kessy et~al.(2018)Kessy, Lewin, and Strimmer]{kessy_2018_optimal}
A.~Kessy, A.~Lewin, and K.~Strimmer, ``Optimal whitening and decorrelation,''
  \emph{The American Statistician}, vol.~72, no.~4, pp. 309--314, 2018.

\bibitem[Li et~al.(2018)Li, Liu, Li, Yang, and Kautz]{li_2018_closed}
Y.~Li, M.-Y. Liu, X.~Li, M.-H. Yang, and J.~Kautz, ``A closed-form solution to
  photorealistic image stylization,'' in \emph{Proceedings of the European
  Conference on Computer Vision (ECCV)}, 2018, pp. 453--468.

\bibitem[Qiao et~al.(2021)Qiao, Cui, Huang, Liu, Bao, and
  Li]{qiao_2021_efficient}
Y.~Qiao, J.~Cui, F.~Huang, H.~Liu, C.~Bao, and X.~Li, ``Efficient style-corpus
  constrained learning for photorealistic style transfer,'' \emph{IEEE
  Transactions on Image Processing}, vol.~30, pp. 3154--3166, 2021.

\bibitem[Wang et~al.(2023)Wang, Li, Zhang, and Feng]{wang_2023_rethinking}
Q.~Wang, S.~Li, X.~Zhang, and G.~Feng, ``Rethinking neural style transfer:
  Generating personalized and watermarked stylized images,'' in
  \emph{Proceedings of the 31st ACM International Conference on Multimedia},
  2023, pp. 6928--6937.

\bibitem[Kang et~al.(2007)Kang, Lee, and Chui]{kang_2007_coherent}
H.~Kang, S.~Lee, and C.~K. Chui, ``Coherent line drawing,'' in
  \emph{Proceedings of the 5th international symposium on Non-photorealistic
  animation and rendering}, 2007, pp. 43--50.

\bibitem[Mallat(1999)]{mallat_1999_wavelet}
S.~Mallat, \emph{A Wavelet Tour of Signal Processing}.\hskip 1em plus 0.5em
  minus 0.4em\relax Elsevier, 1999.

\bibitem[Nason and Silverman(1995)]{nason_1995_stationary}
G.~P. Nason and B.~W. Silverman, ``The stationary wavelet transform and some
  statistical applications,'' in \emph{Wavelets and statistics}.\hskip 1em plus
  0.5em minus 0.4em\relax Springer, 1995, pp. 281--299.

\bibitem[Lu and Do(2003)]{lu_2003_crisp}
Y.~Lu and M.~N. Do, ``Crisp contourlets: a critically sampled directional
  multiresolution image representation,'' in \emph{Wavelets: Applications in
  Signal and Image Processing X}, vol. 5207.\hskip 1em plus 0.5em minus
  0.4em\relax SPIE, 2003, pp. 655--665.

\bibitem[Burt and Adelson(1987)]{burt_1987_laplacian}
P.~J. Burt and E.~H. Adelson, ``The {Laplacian} pyramid as a compact image
  code,'' in \emph{Readings in computer vision}.\hskip 1em plus 0.5em minus
  0.4em\relax Elsevier, 1987, pp. 671--679.

\bibitem[Lu and Do(2007)]{lu_2007_multidimensional}
Y.~M. Lu and M.~N. Do, ``Multidimensional directional filter banks and
  surfacelets,'' \emph{IEEE Transactions on Image Processing}, vol.~16, no.~4,
  pp. 918--931, 2007.

\bibitem[Villani(2003)]{villani_2003_topics}
C.~Villani, \emph{Topics in Optimal Transportation}, ser. Graduate Studies in
  Mathematics.\hskip 1em plus 0.5em minus 0.4em\relax American Mathematical
  Society, 2003.

\bibitem[Takatsu(2010)]{takatsu_2010_wasserstein}
A.~Takatsu, ``On wasserstein geometry of gaussian measures,''
  \emph{Probabilistic Approach to Geometry}, vol.~57, pp. 463--472, 2010.

\bibitem[Bhatia et~al.(2019)Bhatia, Jain, and Lim]{bhatia_2019_bures}
R.~Bhatia, T.~Jain, and Y.~Lim, ``On the bures--wasserstein distance between
  positive definite matrices,'' \emph{Expositiones Mathematicae}, vol.~37,
  no.~2, pp. 165--191, 2019.

\bibitem[Mroueh(2020)]{mroueh_2020_wasserstein}
Y.~Mroueh, ``Wasserstein style transfer,'' in \emph{Proceedings of the
  Conference on Artificial Intelligence and Statistics}, ser. Proceedings of
  Machine Learning Research, vol. 108, 2020, pp. 842--852.

\bibitem[Delon and Desolneux(2020)]{delon_2020_wasserstein}
J.~Delon and A.~Desolneux, ``A wasserstein-type distance in the space of
  gaussian mixture models,'' \emph{SIAM Journal on Imaging Sciences}, vol.~13,
  no.~2, pp. 936--970, 2020.

\bibitem[Do and Vetterli(2002{\natexlab{b}})]{do_2002_rotation}
M.~N. Do and M.~Vetterli, ``Rotation invariant texture characterization and
  retrieval using steerable wavelet-domain hidden markov models,'' \emph{IEEE
  Transactions on Multimedia}, vol.~4, no.~4, pp. 517--527, 2002.

\bibitem[Kang et~al.(2008)Kang, Lee, and Chui]{kang_2008_flow}
H.~Kang, S.~Lee, and C.~K. Chui, ``Flow-based image abstraction,'' \emph{IEEE
  Transactions on Visualization and Computer Graphics}, vol.~15, no.~1, pp.
  62--76, 2008.

\bibitem[Wang et~al.(2004)Wang, Bovik, Sheikh, and Simoncelli]{wang_2004_image}
Z.~Wang, A.~C. Bovik, H.~R. Sheikh, and E.~P. Simoncelli, ``Image quality
  assessment: from error visibility to structural similarity,'' \emph{IEEE
  Transactions on Image Processing}, vol.~13, no.~4, pp. 600--612, 2004.

\bibitem[Zhang et~al.(2018)Zhang, Isola, Efros, Shechtman, and
  Wang]{zhang_2018_unreasonable}
R.~Zhang, P.~Isola, A.~A. Efros, E.~Shechtman, and O.~Wang, ``The unreasonable
  effectiveness of deep features as a perceptual metric,'' in \emph{Proceedings
  of the IEEE Conference on Computer Vision and Pattern Recognition}, 2018, pp.
  586--595.

\bibitem[Heusel et~al.(2017)Heusel, Ramsauer, Unterthiner, Nessler, and
  Hochreiter]{heusel_2017_gans}
M.~Heusel, H.~Ramsauer, T.~Unterthiner, B.~Nessler, and S.~Hochreiter, ``{GAN}s
  trained by a two time-scale update rule converge to a local {Nash}
  equilibrium,'' \emph{Advances in Neural Information Processing Systems},
  vol.~30, 2017.

\end{thebibliography}

\end{document}